\def\year{2021}\relax
\documentclass[letterpaper]{article} 
\usepackage{aaai21}  
\usepackage{times}  
\usepackage{helvet} 
\usepackage{courier}  
\usepackage[hyphens]{url}  
\usepackage{graphicx} 
\usepackage{natbib}  
\usepackage{caption} 
\title{Group Fairness by Probabilistic Modeling with Latent Fair Decisions}
\author{
    YooJung Choi, Meihua Dang, and Guy Van den Broeck \\
}
\affiliations{
    Computer Science Department\\
    University of California, Los Angeles \\
    \{yjchoi,mhdang,guyvdb\}@cs.ucla.edu
}

\usepackage{booktabs}       
\usepackage{amsfonts}       
\usepackage{nicefrac}       
\usepackage{microtype}      

\usepackage{color,soul,booktabs}
\usepackage{stmaryrd}
\usepackage[usenames,dvipsnames]{xcolor}
\usepackage{tikz}
\usepackage{xspace}
\usepackage{graphicx}
\usepackage{amsfonts}
\usepackage{amsmath}
\usepackage{amsthm}
\usepackage{amssymb}
\usepackage{mathtools}
\usepackage{subcaption}
\usepackage{dsfont}
\usepackage[linesnumbered,lined,boxed,commentsnumbered,ruled,vlined,noend]{algorithm2e}
\usepackage{multirow}

\DeclareMathOperator{\EF}{EF}
\DeclareMathOperator{\pr}{Pr}

\usetikzlibrary{trees}
\usetikzlibrary{arrows,shapes,shapes.geometric,backgrounds}
\usetikzlibrary{calc}
\usetikzlibrary{decorations.markings}
\usetikzlibrary{circuits.logic.US}
\usetikzlibrary{fit}
\usetikzlibrary{spn}
\usetikzlibrary{positioning}

\tikzstyle{bnarrow}=[
    decoration={markings,mark=at position 1 with {\arrow[scale=1.5]{>}}},
    postaction={decorate},
    shorten >=0.7pt,
    >=latex,
    line width=0.3
]
\tikzstyle{bayesnet}=[
  >=latex, thick, auto
]
\tikzstyle{bnnode}=[
  draw,ellipse,minimum size=7mm,inner sep=1pt,font=\small
]
\tikzstyle{tnode}=[
  font=\small
]

\newcommand{\midlinewidth}{1.0pt}
\newcommand{\middist}{17pt}
\newcommand{\halfdist}{13pt}

\DeclareMathOperator{\Ex}{\mathbb{E}}

\newcommand{\given}{\vert}

\newcommand{\rvars}[1]{\ensuremath{\mathbf{#1}}\xspace}
\newcommand{\Xs}{\rvars{X}}

\newcommand{\Zs}{\rvars{Z}}

\newcommand{\Ss}{\rvars{S}}

\newcommand{\Es}{\rvars{E}}

\newcommand{\jstate}[1]{\ensuremath{\mathbf{#1}}\xspace}
\newcommand{\xs}{\jstate{x}}

\newcommand{\zs}{\jstate{z}}

\newcommand{\es}{\jstate{e}}

\newcommand{\D}{\ensuremath{\mathcal{D}}}

\newcommand{\PC}{\ensuremath{\mathcal{C}}}
\newcommand{\Graph}{\ensuremath{\mathcal{G}}}
\newcommand{\paras}{\mathcal{\boldsymbol{\theta}}}

\newcommand\independent{\protect\mathpalette{\protect\independenT}{\perp}}
\def\independenT#1#2{\mathrel{\rlap{$#1#2$}\mkern2mu{#1#2}}}
\newcommand{\eat}[1]{}

\newcommand{\strudel}{\textsc{Strudel}}
\newcommand{\fairpc}{\textsc{FairPC}}
\newcommand{\twonb}{\textsc{2NB}}
\newcommand{\latentnb}{\textsc{LatNB}}
\newcommand{\nlatent}{\textsc{NLatPC}}
\newcommand{\fairreduct}{\textsc{Reduction}}
\newcommand{\fairclass}{\textsc{FairLR}}
\newcommand{\reweight}{\textsc{Reweight}}
\newcommand{\rand}{\textsc{Rand}}
\newcommand{\lr}{\textsc{LR}}

\newcommand{\ch}{\ensuremath{\mathsf{ch}}}
\newcommand{\pa}{\ensuremath{\mathsf{pa}}}

\newcommand{\supp}{\ensuremath{\mathsf{supp}}}

\theoremstyle{definition}

\newtheorem{prop}{Proposition}
\newtheorem{defn}{Definition}

\pgfplotsset{compat=newest,compat/show suggested version=false}

\begin{document}
\maketitle
\begin{abstract}
    Machine learning systems are increasingly being used to make impactful decisions such as loan applications and criminal justice risk assessments, and as such, ensuring fairness of these systems is critical. This is often challenging as the labels in the data are biased. This paper studies learning fair probability distributions from biased data by explicitly modeling a latent variable that represents a hidden, unbiased label. In particular, we aim to achieve demographic parity by enforcing certain independencies in the learned model. We also show that group fairness guarantees are meaningful only if the distribution used to provide those guarantees indeed captures the real-world data. In order to closely model the data distribution, we employ probabilistic circuits, an expressive and tractable probabilistic model, and propose an algorithm to learn them from incomplete data. We evaluate our approach on a synthetic dataset in which observed labels indeed come from fair labels but with added bias, and demonstrate that the fair labels are successfully retrieved. Moreover, we show on real-world datasets that our approach not only is a better model than existing methods of how the data was generated but also achieves competitive accuracy.
\end{abstract}

\section{Introduction}
\label{sec:intro}

As machine learning algorithms are being increasingly used in real-world decision making scenarios, there has been growing concern that these methods may produce decisions that discriminate against particular groups of people. The relevant applications include online advertising, hiring, loan approvals, and criminal risk assessment~\cite{datta2015automated,barocas2016big,chouldechova2017fair,berk2018fairness}.
To address these concerns, various methods have been proposed to quantify and ensure fairness in automated decision making systems~\cite{chouldechova2017fair,dwork2012fairness,feldman2015certifying,kusner2017counterfactual,kamishima2012fairness,zemel2013learning}.
A widely used notion of fairness is demographic parity, which states that sensitive attributes such as gender or race must be statistically independent of the class predictions.

In this paper, we study the problem of enforcing demographic parity in probabilistic classifiers. In particular, we focus on the fact that class labels in the data are often biased, and then propose a latent variable approach that treats the observed labels as biased proxies of hidden, fair labels that satisfy demographic parity.
The process that generated bias is modeled by a probability distribution over the fair label, observed label, and other features including the sensitive attributes.
Moreover, we show that group fairness guarantees for a probabilistic model hold in the real world only if the model accurately captures the real-world data.
Therefore, the goal of learning a fair probabilistic classifier also entails learning a distribution that achieves high likelihood.

Our first contribution is to systematically derive the assumptions of a fair probabilistic model in terms of independence constraints. Each constraint serves the purpose of explaining how the observed, biased labels come from hidden fair labels and/or ensuring that the model closely represents the data distribution.
Secondly, we propose an algorithm to learn probabilistic circuits (PCs)~\cite{pcTutorialUAI}, a type of tractable probabilistic models, so that the fairness constraints are satisfied. 
Specifically, this involves encoding independence assumptions into the circuits and developing an algorithm to learn PCs from incomplete data, as we have a latent variable.
Finally, we evaluate our approach empirically on synthetic and real-world datasets, comparing against existing fair learning methods as well as a baseline we propose that does not include a latent variable. The experiments demonstrate that our method achieves high likelihoods that indeed translate to more trustworthy fairness guarantees.
It also has high accuracy for predicting the true fair labels in the synthetic data, and the predicted fair decisions can still be close to unfair labels in real-world data.

\section{Related Work}
\label{sec:related}

Several frameworks have been proposed to design fairness-aware systems. We discuss a few of them here and refer to \citet{romei2014multidisciplinary,barocas-hardt-narayanan} for a more comprehensive review.

Some of the most prominent fairness frameworks include individual fairness and group fairness.
Individual fairness~\citep{dwork2012fairness} is based on the idea that similar individuals should receive similar treatments, although defining similarity between individuals can be challenging.
On the other hand, group fairness aims to equalize some statistics across groups defined by sensitive attributes. These include equality of opportunity~\citep{hardt2016equality} and demographic (statistical) parity~\citep{calders2010three,kamiran2009classifying} as well as its relaxed notion of disparate impact~\citep{feldman2015certifying,zafar2017fairness}.

There are several approaches to achieve group fairness, which can be broadly categorized into (1) pre-processing data to remove bias~\citep{zemel2013learning,kamiran2009classifying,calmon2017optimized}, (2) post-processing of model outputs such as calibration and threshold selection~\citep{hardt2016equality,pleiss2017fairness}, and (3) in-processing which incorporates fairness constraints directly in learning/optimization~\citep{corbett2017algorithmic,agarwal2018reductions,kearns2018preventing}.
Some recent works on group fairness also consider bias in the observed labels, both for evaluation and learning~\cite{Fogliato2020FairnessEI,blum2020recovering,jiang2020identifying}. For instance, \citet{blum2020recovering} studies empirical risk minimization (ERM) with various group fairness constraints and showed that ERM constrained by demographic parity does not recover the Bayes optimal classifier under one-sided, single-group label noise (this setting is subsumed by ours).
In addition, \citet{jiang2020identifying} developed a pre-processing method to learn fair classifiers under noisy labels, by reweighting according to an unknown, fair labeling function. Here, the observed labels are assumed to come from a biased labeling function that is the ``closest'' to the fair one; whereas, we aim to find the bias mechanism that best explains the observed data.

We would like to point out that while pre-processing methods have the advantage of allowing any model to be learned on top of the processed data, it is also known that certain modeling assumptions can result in bias even when learning from fair data~\citep{ChoiAAAI20}. Moreover, certain post-processing methods to achieve group fairness are shown to be suboptimal under some conditions~\citep{woodworth2017learning}.
Instead, we take the in-processing approach to explicitly optimize the model's performance while enforcing fairness.

Many fair learning methods make use of probabilistic models such as Bayesian networks~\citep{calders2010three,mancuhan2014combating}. Among those, perhaps the most related to our approach is the latent variable naive Bayes model by \citet{calders2010three}, which also assumes a latent decision variable to make fair predictions. However, they make a naive Bayes assumption among features.
We relax this assumption and will later demonstrate how this helps in more closely modeling the data distribution, as well as providing better fairness guarantees.

\section{Latent Fair Decisions}
\label{sec:model}

We use uppercase letters (e.g., $X$) for discrete random variables (RVs) and lowercase letters ($x$) for their assignments. Negation of a binary assignment $x$ is denoted by $\bar{x}$. Sets of RVs are denoted by bold uppercase letters ($\Xs$), and their joint assignments by bold lowercase ($\xs$). 
%
%
Let $S$ denote a \textit{sensitive attribute}, such as gender or race, and let $\Xs$ be the \textit{non-sensitive attributes} or features. In this paper, we assume $S$ is a binary variable for simplicity, but our method can be easily generalized to multiple multi-valued sensitive attributes.
We have a dataset $\D$ in which each individual is characterized by variables $S$ and $\Xs$ and labeled with a binary decision/class variable $D$.

One of the most popular and yet simple fairness notions is demographic (or statistical) parity. It requires that the classification is independent of the sensitive attributes; i.e., the rate of positive classification is the same across groups defined by the sensitive attributes.
Since we focus on probabilistic classifiers, we consider a generalized version introduced by \citet{pleiss2017fairness}, sometimes also called \emph{strong demographic parity}~\citep{jiang2019wasserstein}:
\begin{defn}[Generalized demographic parity]
    Suppose $f$ is a probabilistic classifier and $p$ is a distribution over variables $\Xs$ and $S$. Then $f$ satisfies demographic parity w.r.t.\ $p$ if: 
    \begin{equation*}
        \Ex_p[f(\Xs,S) \mid S=1) = \Ex_p[f(\Xs,S) \mid S=0].
    \end{equation*}
\end{defn}
Probabilistic classifiers are often obtained from joint distributions $\Pr(.)$ over $D,\Xs,S$ by computing $\Pr(D \given \Xs,S)$. Then we say the distribution satisfies demographic parity if $\Pr(D \given S\!=\!1) = \Pr(D \given S\!=\!0)$, i.e., $D$ is independent of $S$.

\subsection{Motivation}

A common fairness concern when learning decision making systems is that the dataset used is often biased.
In particular, observed labels may not be the true target variable but only its proxy. For example, re-arrest is generally used as a label for recidivism prediction, but it is not equivalent to recidivism and may be biased.
We will later show how the relationship between observed label and true target can be modeled probabilistically using a latent variable.

Moreover, probabilistic group fairness guarantees hold in the real world only if the model accurately captures the real world distribution. In other words, using a model that only achieves low likelihood w.r.t\ the data, it is easy to give false guarantees.
For instance, consider a probabilistic classifier $f(X,S)$ over a binary sensitive attribute $S$ and non-sensitive attribute $X$ shown below. 
\begin{center}
\scalebox{0.76}{
    \begin{tabular}{lc|cc|cc}
    \toprule
    $S,X$ & $f(X,S)$ & $P_{\text{data}}(X\given S)$ & $\Ex_{P_{\text{data}}}[f\given S]$ & $Q(X\given S)$ & $\Ex_{Q}[f\given S]$ \\
    \midrule
    1,1 & 0.8 & 0.7 & \multirow{2}{*}{0.65} &   0.5 & \multirow{2}{*}{0.55} \\
    1,0 & 0.3 & 0.3 & &                         0.5 & \\ 
    \midrule
    0,1 & 0.7 & 0.4 & \multirow{2}{*}{0.52} &   0.5 & \multirow{2}{*}{0.55}\\
    0,0 & 0.4 & 0.6 & &                         0.5\\
    \bottomrule
\end{tabular}
}
\end{center}
Suppose in the data, the probability of $X=1$ given $S=1$ (resp.\ $S=0$) is 0.7 (resp.\ 0.4). Then this classifier does not satisfy demographic parity, as the expected prediction for group $S=1$ is $0.8\cdot0.7+0.3\cdot0.3=0.65$ while for group $S=0$ it is 0.52.
On the other hand, suppose you have a distribution $Q$ that incorrectly assumes the feature $X$ to be uniform and independent of $S$. Then you would conclude, incorrectly, that the prediction is indeed fair, with the average prediction for both protected groups being 0.55. Therefore, to provide meaningful fairness guarantees, we need to model the data distribution closely, i.e., with high likelihood.

\subsection{Modeling with a latent fair decision}
We now describe our proposed latent variable approach to address the aforementioned issues.
We suppose there is a hidden variable that represents the true label without discrimination. This latent variable is denoted as $D_f$ and used for prediction instead of $D$; i.e., decisions for future instances can be made by inferring the conditional probability $\Pr(D_f \given \es)$ given some feature observations $\es$ for $\Es \subseteq \Xs \cup S$. 
We assume that the latent variable $D_f$ is independent of $S$, thereby satisfying demographic parity.
Moreover, the observed label $D$ is modeled as being generated from the fair label by altering its values with different probabilities depending on the sensitive attribute. In other words, the probability of $D$ being positive depends on both $D_f$ and $S$.

\begin{figure}[tb]
    \centering
    \begin{subfigure}[b]{0.49\columnwidth}
        \centering
    \scalebox{0.9}{
      \begin{tikzpicture}[bayesnet]
          
      \def\lone{0bp}
      \def\ltwo{-40bp}
    
      \node (S) at (-20bp,\lone) [bnnode] {$S$};
      \node (Df) at (20bp,\lone) [bnnode] {$D_f$};
      \node (X) at (-20bp,\ltwo) [bnnode] {$\Xs$};
      \node (D) at (20bp,\ltwo) [bnnode] {$D$};
    
      \begin{scope}[on background layer]
        \draw [bnarrow] (S) -- (X);
        \draw [bnarrow] (S) -- (D);
        \draw [bnarrow] (Df) -- (X);
        \draw [bnarrow] (Df) -- (D);
      \end{scope}
      \end{tikzpicture}}
      \caption{}
      \label{fig:fair-BN}
    \end{subfigure}
    \begin{subfigure}[b]{0.49\columnwidth}
        \centering
    \scalebox{0.9}{
      \begin{tikzpicture}[bayesnet]
          
      \def\lone{0bp}
      \def\ltwo{-40bp}
    
      \node (S) at (-20bp,\lone) [bnnode] {$S$};
      \node (D) at (20bp,\lone) [bnnode] {$D$};
      \node (X) at (-20bp,\ltwo) [bnnode] {$\Xs$};
      
      \begin{scope}[on background layer]
        \draw [bnarrow] (S) -- (X);
        \draw [bnarrow] (D) -- (X);
      \end{scope}
      \end{tikzpicture}}
      \caption{}
      \label{fig:alt-BN}
    \end{subfigure}
    \caption{Bayesian network structures that represent the proposed fair latent variable approach (left) and model without a latent variable (right). Abusing notation, the set of features $\Xs$ is represented as a single node, but refers to some local Bayesian network over $\Xs$.}
\end{figure}
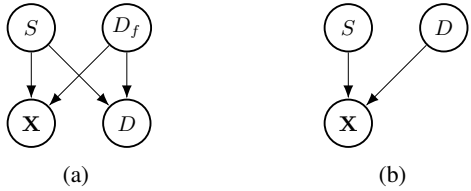

In addition, our model also assumes that the observed label $D$ and non-sensitive features $\Xs$ are conditionally independent given the latent fair decision and sensitive attributes, i.e., $D \independent \Xs \given D_f,S$. 
This is a crucial assumption to learn the model from data where $D_f$ is hidden. To illustrate why, suppose there is no such independence.
Then the induced model allows variables $S, \Xs, D$ to depend on one another freely. Thus, such model can represent any marginal distribution over these variables, regardless of the parameters for $D_f$. 
We can quickly see this from the fact that for all $s,\xs,d$,
\begin{align*}
    \Pr(s\xs d) &= \Pr(s) \Pr(\xs d \given s)  \\
    &= \Pr(s) \big( \Pr(\xs d \given s,D_f\!=\!1) \Pr(D_f\!=\!1) \\
    &\phantom{=\Pr(s) \big(}+ \Pr(\xs d \given s, D_f\!=\!0) \Pr(D_f\!=\!0) \big).
\end{align*}
That is, multiple conditional distributions involving the latent fair decision $D_f$ will result in the same marginal distribution over $S,\Xs,D$, and thus the real joint distribution is not identifiable when learning from data where $D_f$ is completely hidden.
For instance, the learner will not be incentivized to represent any dependence between $D_f$ and other features, and may return a model in which the latent decision variable is completely independent of the observed variables.
This is clearly undesirable because we want to use the latent variable to make decisions based on feature observations.

The independence assumptions of our proposed model are summarized as a Bayesian network structure in Figure~\ref{fig:fair-BN}.
Note that the set of features $\Xs$ is represented as a single node, as we do not make any independence assumptions among the features. In practice, we learn the statistical relationships between these variables from data.
This is in contrast to the latent variable model in \citet{calders2010three} which had a naive Bayes assumption among the non-sensitive features.
As we will later show empirically, such strong assumption not only affects the prediction quality but also limits the fairness guarantee, as it will hold only if the naive Bayes assumption is indeed true in the data distribution.

The latent variable not only encodes the intuition that observed labels may be biased, but it also has advantages in achieving high likelihood with respect to data. Consider an alternative way to satisfy statistical parity: by directly enforcing independence between the observed decision variable $D$ and sensitive attributes $\Ss$: see Figure~\ref{fig:alt-BN}. 
%
We will show that, on the same data, our proposed model can always achieve marginal likelihood at least as high as the model without a latent decision variable. We can enforce the independence of $D$ and $\Ss$ by setting the latent variable $D_f$ to always be equal to $D$, which results in a marginal distribution over $S,\Xs,D$ with the same independencies as in Figure~\ref{fig:alt-BN}:
\begin{align*}
    &\Pr(s\xs d) \\
    &= \Pr(\xs \mid s,D_f\!=\!1) \Pr(d \mid s,D_f\!=\!1)\Pr(s)\Pr(D_f\!=\!1) \\
    &\quad + \Pr(\xs \mid s,D_f\!=\!0)\Pr(d \mid s, D_f\!=\!0)\Pr(s)\Pr(D_f\!=\!0) \\
    &= \Pr(\xs \mid s d) \Pr(s) \Pr(d)
\end{align*}
Thus, any fair distribution without the latent decision can also be represented by our latent variable approach.
In addition, our approach will achieve strictly better likelihood if the observed data does not satisfy demographic parity, because it can also model distributions where $D$ and $S$ are dependent.



Lastly, we emphasize that Bayesian network structures were used in this section only to illustrate the independence assumptions of our model.
In practice, other probabilistic models can be used to represent the distribution as long as they satisfy our independence assumptions; we use probabilistic circuits as discussed in the next section.

\section{Learning Fair Probabilistic Circuits}
\label{sec:pc}

There are several challenges in modeling a fair probability distribution.
First, as shown previously, fairness guarantees hold with respect to the modeled distribution, and thus we want to closely model the data distribution. A possible approach is to learn a deep generative model such as a generative adversarial networks (GANs)~\citep{goodfellow2014generative}. However, then we must resort to approximate inference, or deal with models that have no explicit likelihood, and the fairness guarantees no longer hold.
An alternative is to use models that allow exact inference such as Bayesian networks. Unfortunately, marginal inference, which is needed to make predictions $\Pr(D_f \given \es)$, is \#P-hard for general BNs~\cite{Roth1996}. Tree-like BNs such as naive Bayes allow polytime inference, but they are not expressive enough to accurately capture the real world distribution.
Hence, the second challenge is to also support tractable exact inference without sacrificing expressiveness.
%
Lastly, the probabilistic modeling method we choose must be able to encode the independencies outlined in the previous section, to satisfy demographic parity and to learn a meaningful relationship between the latent fair decision and other variables.
In the following, we give some background on \emph{probabilistic circuits (PCs)} and show how they satisfy each of the above criteria. Then we will describe our proposed algorithm to learn fair probabilistic circuits from data.

\subsection{Probabilistic Circuits}
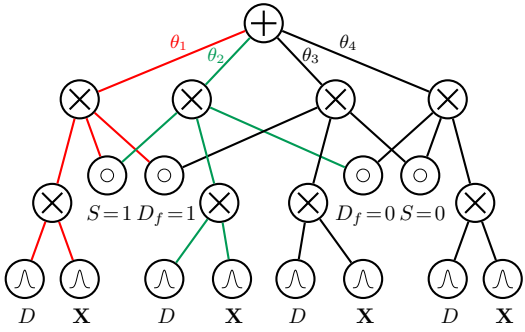
\begin{figure}[tb]
    \centering
    \scalebox{0.8}{
    \begin{tikzpicture}
    	\sumnode[line width=\midlinewidth]{s1};
    	\prodnode[line width=\midlinewidth, below=\middist of s1, xshift=-2*\middist]{p2};
    	\prodnode[line width=\midlinewidth, below=\middist of s1, xshift=2*\middist]{p3};
        \prodnode[line width=\midlinewidth, left=2*\middist of p2]{p4};
        \prodnode[line width=\midlinewidth, right=2*\middist of p3]{p5};
    
    	\bernode[line width=\midlinewidth, below=\middist of p4, xshift=\halfdist]{v1}{$S\!=\!1$};
        \bernode[line width=\midlinewidth, below=\middist of p2, xshift=-\halfdist]{v2}{$D_f\!=\!1$};
    	\bernode[line width=\midlinewidth, below=\middist of p3, xshift=\halfdist]{v3}{$D_f\!=\!0$};
        \bernode[line width=\midlinewidth, below=\middist of p5, xshift=-\halfdist]{v4}{$S\!=\!0$};
        
        \prodnode[line width=\midlinewidth, below=\middist+\halfdist of p4, xshift=-\halfdist]{p6};
        \prodnode[line width=\midlinewidth, below=\middist+\halfdist of p2, xshift=\halfdist]{p7};
        \prodnode[line width=\midlinewidth, below=\middist+\halfdist of p3, xshift=-\halfdist]{p8};
        \prodnode[line width=\midlinewidth, below=\middist+\halfdist of p5, xshift=\halfdist]{p9};
        
    	\contnode[line width=\midlinewidth, below=\middist of p6, xshift=-\halfdist]{v5}{$D$};
    	\contnode[line width=\midlinewidth, below=\middist of p6, xshift=\halfdist]{v6}{$\Xs$};
    	\contnode[line width=\midlinewidth, below=\middist+\halfdist of v2]{v7}{$D$};
    	\contnode[line width=\midlinewidth, right=\halfdist of v7]{v8}{$\Xs$};
    	\contnode[line width=\midlinewidth, below=\middist+\halfdist of v3]{v9}{$\Xs$};
    	\contnode[line width=\midlinewidth, left=\halfdist of v9]{v10}{$D$};
    	\contnode[line width=\midlinewidth, below=\middist of p9, xshift=-\halfdist]{v11}{$D$};
    	\contnode[line width=\midlinewidth, below=\middist of p9, xshift=\halfdist]{v12}{$\Xs$};
    	
      \weigedge[line width=\midlinewidth,above,pos=0.45,color=red] {s1} {p4} {\footnotesize $\color{red}\theta_1$};
      \weigedge[line width=\midlinewidth,left,pos=0.37,color=ForestGreen] {s1} {p2} {\footnotesize $\color{ForestGreen}\theta_2$};
      \weigedge[line width=\midlinewidth,right,pos=0.37] {s1} {p3} {\footnotesize $\theta_3$};
      \weigedge[line width=\midlinewidth,above,pos=0.45] {s1} {p5} {\footnotesize $\theta_4$};
    
      \edge[line width=\midlinewidth,left,>=stealth,draw=red] {p4} {v1};
      \edge[line width=\midlinewidth,right,>=stealth,draw=red] {p4} {v2};
      \edge[line width=\midlinewidth,left,>=stealth,color=ForestGreen] {p2} {v1};
      \edge[line width=\midlinewidth,right,>=stealth,color=ForestGreen] {p2} {v3};
      \edge[line width=\midlinewidth,left,>=stealth] {p3} {v2};
      \edge[line width=\midlinewidth,right,>=stealth] {p3} {v4};
      \edge[line width=\midlinewidth,left,>=stealth] {p5} {v3};
      \edge[line width=\midlinewidth,right,>=stealth] {p5} {v4};
      
      \edge[line width=\midlinewidth,left,>=stealth,draw=red] {p4} {p6};
      \edge[line width=\midlinewidth,left,>=stealth,color=ForestGreen] {p2} {p7};
      \edge[line width=\midlinewidth,left,>=stealth] {p3} {p8};
      \edge[line width=\midlinewidth,left,>=stealth] {p5} {p9};
      
      \edge[line width=\midlinewidth,left,>=stealth,draw=red] {p6} {v5};
      \edge[line width=\midlinewidth,left,>=stealth,draw=red] {p6} {v6};
      \edge[line width=\midlinewidth,left,>=stealth,color=ForestGreen] {p7} {v7};
      \edge[line width=\midlinewidth,left,>=stealth,color=ForestGreen] {p7} {v8};
      \edge[line width=\midlinewidth,left,>=stealth] {p8} {v9};
      \edge[line width=\midlinewidth,left,>=stealth] {p8} {v10};
      \edge[line width=\midlinewidth,left,>=stealth] {p9} {v11};
      \edge[line width=\midlinewidth,left,>=stealth] {p9} {v12};
    \end{tikzpicture}
    }
    \caption{ A probabilistic circuit over variables $S,\Xs,D,D_f$}
    \label{fig:fair-pc}    
\end{figure}

\paragraph{Representation}
\emph{Probabilistic circuits (PCs)}~\cite{PCTuto20} refer to a family of tractable probabilistic models including arithmetic circuits~\cite{darwicheKR02,darwicheJACM-POLY}, sum-product networks~\cite{poon2011sum}, cutset networks~\cite{rahman2014cutset}, and and-or search spaces~\cite{marinescu2005and}. 
A probabilistic circuit $\PC = (\Graph,\paras)$ over RVs $\Xs$ is characterized by its structure $\Graph$ and parameters $\paras$. 
%
The circuit structure $\Graph$ is a directed acyclic graph (DAG) such that each inner node is either a \textit{sum} node or a \textit{product} node, and each \textit{leaf} (input) node is associated with a univariate input distribution. We denote the distribution associated with leaf $n$ by $f_n(.)$. This may be any probability mass function, a special case being an indicator function such as $[X=1]$.
Parameters $\paras$ are each associated with an input edge to a sum node.
Note that a subcircuit rooted at an inner node of a PC is itself a valid PC. Figure~\ref{fig:fair-pc} depicts an example probabilistic circuit.\footnote{The features $\Xs$ and $D$ are shown as leaf nodes for graphical conciseness, but refer to sub-circuits over the respective variables.}
  
Let $\ch(n)$ be the set of children nodes of an inner node $n$. Then a probabilistic circuit $\PC$ over RVs $\Xs$ defines a joint distribution $\Pr_{\PC}(\Xs)$ in a recursive way as follows:
%
\begin{equation*}
{\Pr}_n(\xs)=
\begin{cases}
    f_n(\xs) &\text{if $n$ is a leaf node} \\
    \prod_{c\in\ch(n)} \Pr_c(\xs) &\text{if $n$ is a product node} \\
    \sum_{c\in\ch(n)} \theta_{n,c} \Pr_c(\xs) &\text{if $n$ is a sum node}
\end{cases}
\label{eq:EVI}
\end{equation*}
%
%
Intuitively, a product node $n$ defines a factorized distribution, and a sum node $n$ defines a mixture model parameterized by weights $\{\theta_{n,c}\}_{c\in\ch(n)}$.
%
$\Pr_n$ is also called the \textit{output} of $n$.

%
\paragraph{Properties and inference}
A strength of probabilistic circuits is that (1) they are expressive, achieving high likelihoods on density estimation tasks~\cite{rahman2016merging,LiangUAI17,peharz2020random}, and (2) they support tractable probabilistic inference, enabled by certain structural properties. In particular, PCs support efficient marginal inference if they are smooth and decomposable.
A circuit is said to be \emph{smooth} if for every sum node all of its children depend on the same set of variables; it is \emph{decomposable} if for every product node its children depend on disjoint sets of variables~\citep{darwiche2002knowledge}.
Given a smooth and decomposable probabilistic circuit, computing the marginal probability for any partial evidence is reduced to simply evaluating the circuit bottom-up. This also implies tractable computation of conditional probabilities, which are ratios of marginals. Thus, we can make predictions in time linear in the size of the circuit.

Another useful structural property is \emph{determinism}; a circuit is deterministic if for every complete input $\xs$, at most one child of every sum node has a non-zero output. 
In addition to enabling tractable inference for more queries~\citep{ChoiDarwiche17}, it leads to closed-form parameter estimation of probabilistic circuits given complete data. We also exploit this property for learning PCs with latent variables, which we will later describe in detail.

\paragraph{Encoding independence assumptions}
Next, we demonstrate how we encode the independence assumptions of a fair distribution as in Figure~\ref{fig:fair-BN} in a probabilistic circuit.
Recall the example PC in Figure~\ref{fig:fair-pc}: regardless of parameterization, this circuit structure always encodes a distribution where $D$ is independent of $\Xs$ given $S$ and $D_f$. 
To prove this, we first observe that the four product nodes in the second layer each correspond to four possible assignments to $S$ and $D_f$. For instance, the left-most product node returns a non-zero output only if the input sets both $S=1$ and $D_f=1$.
Effectively, the sub-circuits rooted at these nodes represent conditional distributions $\Pr(D,\Xs \given s, d_f)$ for assignments $s,d_f$. Because the distributions for $D$ and $\Xs$ factorize, we have $\Pr(D,\Xs \given s,d_f) = \Pr(D \given s,d_f) \cdot \Pr(\Xs \given s,d_f)$, thereby satisfying the conditional independence $D \independent \Xs \given D_f,S$.

We also need to encode the independence between $D_f$ and $S$. In the example circuit, each edge parameter $\theta_i$ corresponds to $\Pr(s,d_f)$ for a joint assignment to $S,D_f$; e.g.\ $\theta_1=\Pr(S\!=\!1,D_f\!=\!1)$. With no restriction on these parameters, the circuit structure does not necessarily imply $D_f\!\independent\!S$. Thus, we introduce auxiliary parameters $\phi_s$ and $\phi_{d_f}$ representing $\Pr(S\!=\!1)$ and $\Pr(D_f\!=\!1)$, respectively, and enforce that the circuit parameters for $\Pr(S,D_f)$ factorize as follows:
\begin{gather*}
    \phi_s = \theta_1 + \theta_2, \quad \phi_{d_f} = \theta_1 + \theta_2, \\
    \theta_1 = \phi_s \cdot \phi_{d_f},\; \theta_2 = \phi_s \cdot (1-\phi_{d_f}), \\
    \theta_3 = (1-\phi_s) \cdot \phi_{d_f},\; \theta_4 = (1-\phi_s) \cdot (1-\phi_{d_f}).
\end{gather*}
Hence, when learning these parameters, we limit the degree of freedom such that the four edge parameters are given by two free variables $\phi_s$ and $\phi_{d_f}$ instead of the four $\theta_i$ variables.

Next, we discuss how to learn a fair probabilistic circuit with latent variable from data. This consists of two parts: learning the circuit structure and estimating the parameters of a given structure.
We first study parameter learning in the next section, then structure learning in Section~\ref{sec:str-learning}.

\subsection{Parameter Learning}
\label{sec:param-learn}

Given a complete data set, maximum-likelihood parameters of a smooth, decomposable, and deterministic PC can be computed in closed-form~\citep{KisaVCD14}. 
For an edge between a sum node $n$ and its child $c$, the associated maximum-likelihood parameter for a complete dataset $\D$ is given by:
\begin{equation}
    \theta_{n,c} = F_{\D}(n,c) / \sum_{c\in\ch(n)} F_{\D}(n,c)  \label{eq:param}
\end{equation}
Here, $F_{\D}(n,c)$ is called the \emph{circuit flow} of edge $(n,c)$ given $\D$, and it counts the number of data samples in $\D$ that ``activate'' this edge.
For example, in Figure \ref{fig:fair-pc}, the edges activated by sample $\{D_f\!=\!1,S\!=\!1,d,\xs\}$, for any assignments $d,\xs$, are colored red.\footnote{See Appendix~\ref{sec:app-EF}, available at \url{http://starai.cs.ucla.edu/papers/ChoiAAAI21.pdf}, for a formal definition and proof of Equation~\ref{eq:param}.} 

However, our proposed approach for fair distribution includes a latent variable, and thus must be learned from incomplete data.
One of the most common methods to learn parameters of a probabilistic model from incomplete data is the Expectation Maximization (EM) algorithm~\citep{PGM,darwiche2009}.
EM iteratively completes the data by computing the probability of unobserved values (E-step) and estimates the maximum-likelihood parameters from the expected dataset (M-step).

We now propose an EM parameter learning algorithm for PCs that does not explicitly complete the data, but rather utilizes circuit flows. In particular, we introduce the notion of \emph{expected flows}, which is defined as the following for a given circuit $\PC=(\Graph,\paras)$ over RVs $\Zs$ and an incomplete dataset $\D$:
\begin{align*}
    \EF_{\D,\paras}(n,c) :=& \mathbb{E}_{\pr_{\PC}}[F_{\D_i}(n,c)] \\
    =& \sum_{\D_i \in \D} \sum_{\zs \models \D_i} \pr_{\PC}(\zs \given \D_i) \cdot F_{\zs}(n,c).
\end{align*}
Here, $\D_i$ denotes the i-th sample in the dataset, and $\zs \models \D_i$ are the possible completions of sample $\D_i$. 
For example, in Figure~\ref{fig:fair-pc}, the expected flows of the edges highlighted in red and green, given a sample $\{S\!=\!1,d,\xs\}$, are $\Pr_{\PC}(D_f\!=\!1 \mid S\!=\!1,d,\xs)$ and $\Pr_{\PC}(D_f\!=\!0 \mid S\!=\!1,d,\xs)$, respectively.
Similar to circuit flows, the expected flows for all edges can be computed with a single bottom-up and top-down evaluation of the circuit.
Then, we can perform both the E- and M-step by the following closed-form solution.
\begin{prop}\label{prop:em}
    Given a smooth, decomposable, and deterministic circuit with parameters $\paras$ and an incomplete data $\D$, the parameters for the next EM iteration are given by:
    \begin{equation*}
        \theta_{n,c}^{\text{(new)}} = \EF_{\D,\paras}(n,c) / \sum_{c\in\ch(n)} \EF_{\D,\paras}(n,c).
    \end{equation*}
\end{prop}
Note that this is very similar to the ML estimate from complete data in Eq.\ref{eq:param}, except using expected flows instead of circuit flows. 
Moreover, the expected flow can be computed even if each data sample has different variables missing; thus, the EM method can naturally handle missing values for other features as well.
We refer to Appendix~\ref{sec:app-EF} for details on computing the expected flows and proof for above proposition. 


\paragraph{Initial parameters using prior knowledge}
Typically the EM algorithm is run starting from randomly initialized parameters. 
While the algorithm is guaranteed to improve the likelihood at each iteration until convergence, it still has the problem of multiple local maxima and identifiability, especially when there is a latent variable involved~\citep{PGM}.
Namely, we can converge to different learned models with similar likelihoods but different parameters for the latent fair variable, thus resulting in different behaviors in the prediction task. 
For example, for a given fair distribution, we can flip the value of $D_f$ and the parameters accordingly such that the marginal distribution over $S,\Xs,D$, as well as the likelihood on the dataset, is unchanged. However, this clearly has a significant impact on the predictions which will be completely opposite.

Therefore, instead of random initialization, we encode prior knowledge in the initial parameters that determine $\Pr(D \given S, D_f)$. In particular, it is obvious that $D_f$ should be equal to $D$ if the observed labels are already fair.
Furthermore, for individual predictions, we would want $D_f$ to be close to $D$ as much as possible while ensuring fairness. Thus, we start the EM algorithm from a conditional probability $\Pr(d \given s,d_f) = [d=d_f]$.

\subsection{Structure Learning}
\label{sec:str-learning}
Lastly, we describe how a fair probabilistic circuit structure is learned from data. As described previously, top layers of the circuit are fixed in order to encode the independence assumptions of our latent variable approach.
On the other hand, the sub-circuits over features $\Xs$ can be learned to best fit the data.
We adopt the \strudel\ algorithm to learn the structures~\citep{DangPGM20}.\footnote{PCs learned this way also satisfy properties such as structured decomposability that are not necessary for our use case.}
Starting from a Chow-Liu tree initial distribution~\citep{ChowLiu}, \strudel\ performs a heuristic-based greedy search over possible candidate structures. 
At each iteration, it first selects the edge with the highest circuit flow and the variable with the strongest dependencies on other variables, estimated by the sum of pairwise mutual informations. Then it applies the \textit{split} operation -- a simple structural transformation that ``splits'' the selected edge by introducing new sub-circuits conditioned on the selected variable. 
Intuitively, this operation aims to model the data more closely by capturing the dependence among variables (variable heuristic) appearing in many data samples (edge heuristic).
After learning the structure, we update the parameters of the learned circuit using EM as described previously. 



%
\section{Experiments}
\label{sec:exp}

We now empirically evaluate our proposed model \fairpc\ on real-world benchmark datasets as well as synthetic data.

\paragraph{Baselines}
We first compare \fairpc\ to three other probabilistic methods: fair naive Bayes models (\twonb\ and \latentnb) by \citet{calders2010three} and PCs without latent variable (\nlatent) as described in Section~\ref{sec:model}. 
%
We also compare against existing methods that learn discriminative classifiers satisfying group fairness: (1) \fairclass~\citep{zafar2017fairness}, which learns a classifier subject to co-variance constraints; (2) \fairreduct~\citep{agarwal2018reductions}, which reduces the fair learning problem to cost-sensitive classification problems and learns a randomized classifier subject to fairness constraints; and (3) \reweight~\citep{jiang2020identifying} which corrects bias by re-weighting the data points. All three methods learn logistic regression classifiers, either with constraints or using modified objective functions.

\paragraph{Evaluation criteria} 
For predictive performance, we use accuracy and F1 score. Note that models with latent variables use the latent fair decision $D_f$ to make predictions, while other models directly use $D$.
Moreover, in the real-world datasets, we do not have access to the fair labels and instead evaluate using the observed labels which may be ``noisy'' and biased. We emphasize that the accuracy w.r.t\ unfair labels is not the goal of our method, as we want to predict the true target, not its biased proxy. Rather, it measures how similar the latent variable is to the observed labels, thereby justifying its use as fair decision.
To address this, we also evaluate on synthetic data where fair labels can be generated.

For fairness performance, 
we define the discrimination score as the difference in average prediction probability between the majority and minority groups, i.e., $\Pr(D_f\!=\!1\given S\!=\!0)-\Pr(D_f\!=\!1\given S\!=\!1)$ estimated on the test set.

\begin{figure}[t]
\centering
\begin{tabular}{ccc}
\includegraphics[width=0.28\columnwidth,trim=0cm 1.5cm 3.5cm 0.5cm]{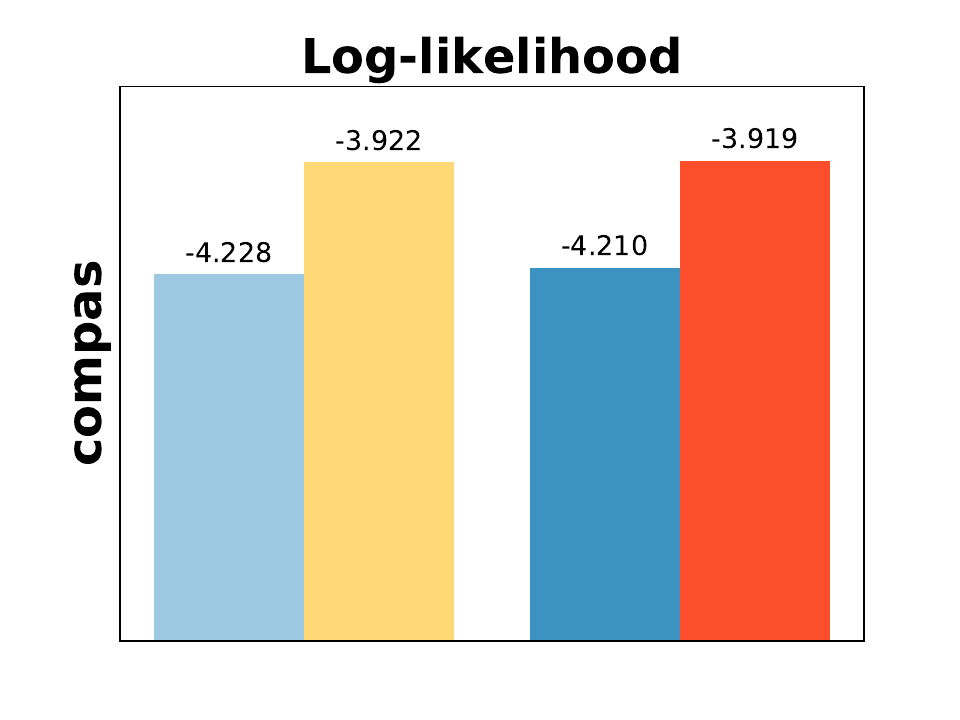}&
\includegraphics[width=0.28\columnwidth,trim=0cm 1.5cm 3.5cm 0.5cm]{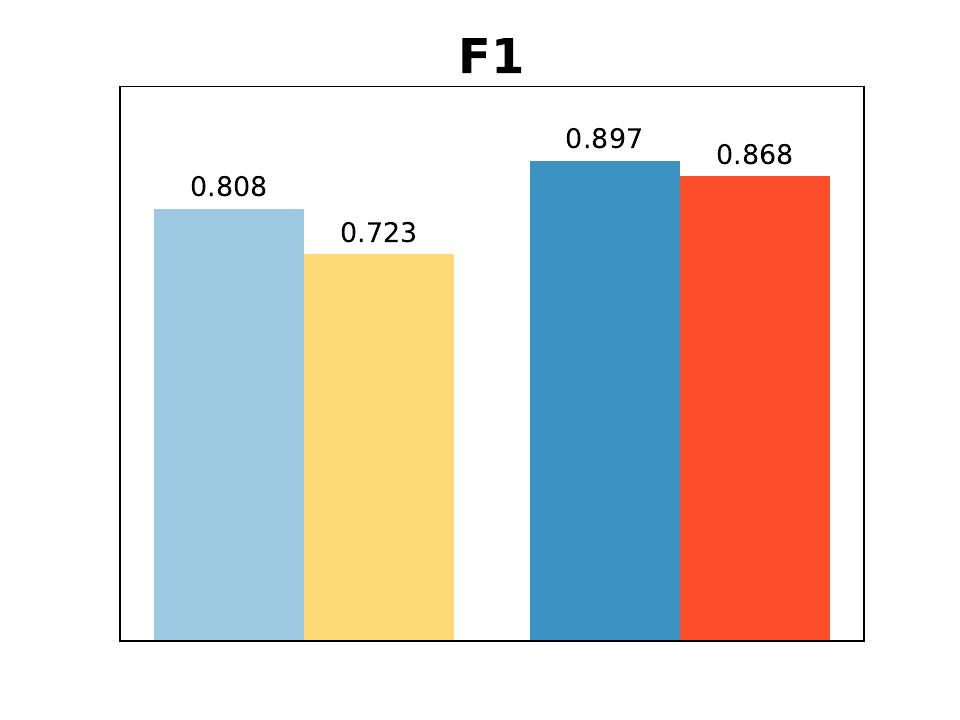}&
\includegraphics[width=0.28\columnwidth,trim=0cm 1.5cm 3.5cm 0.5cm]{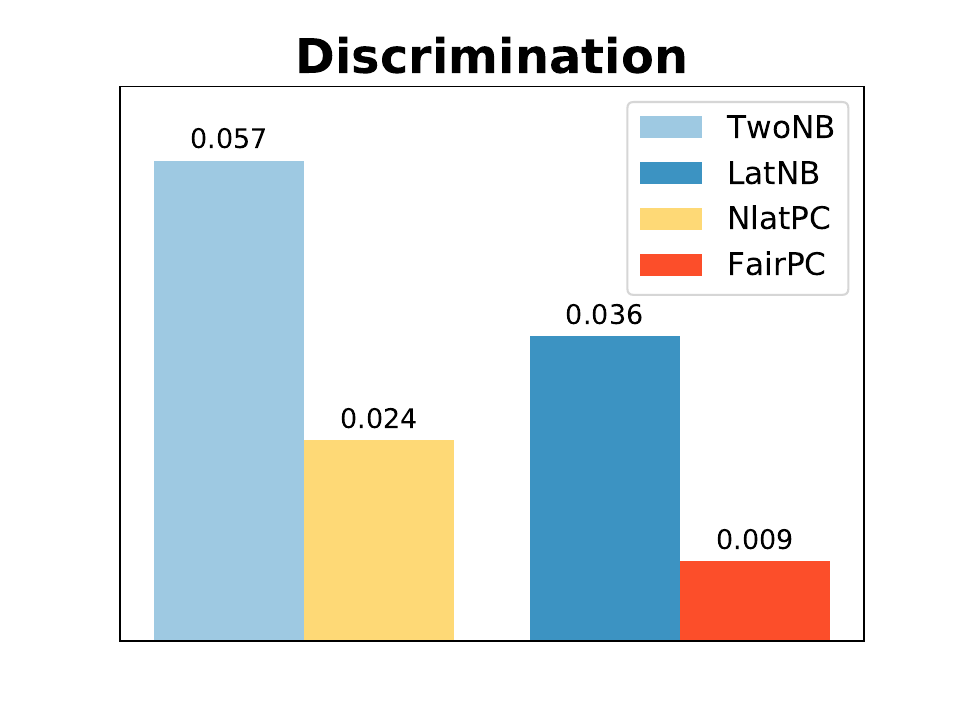}\\
\includegraphics[width=0.28\columnwidth,trim=0cm 1.5cm 3.5cm 0.5cm]{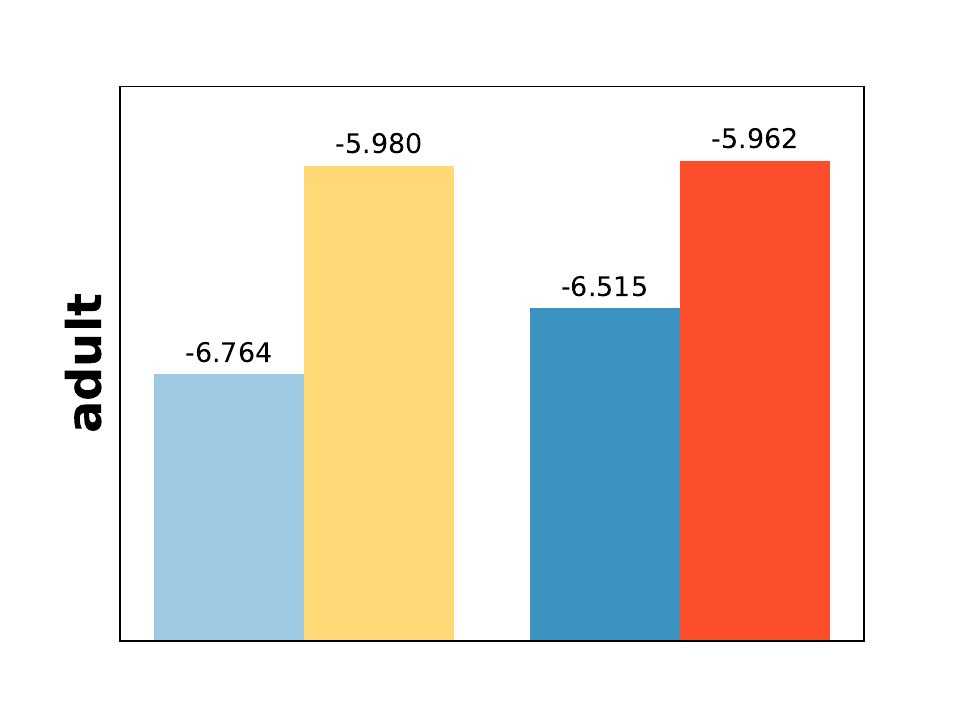}&
\includegraphics[width=0.28\columnwidth,trim=0cm 1.5cm 3.5cm 0.5cm]{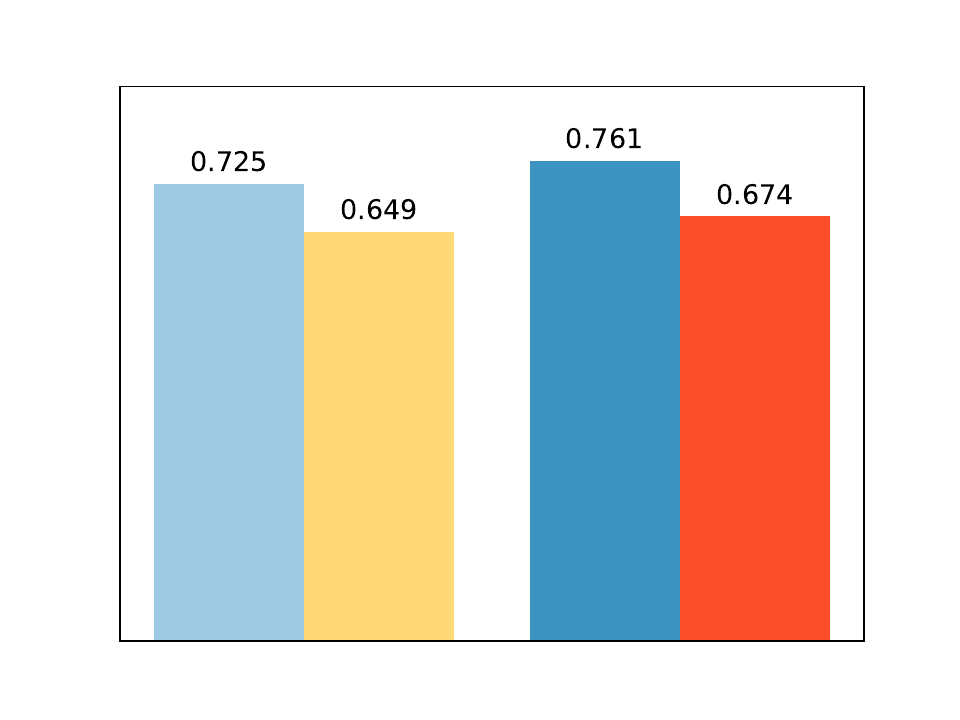}&
\includegraphics[width=0.28\columnwidth,trim=0cm 1.5cm 3.5cm 0.5cm]{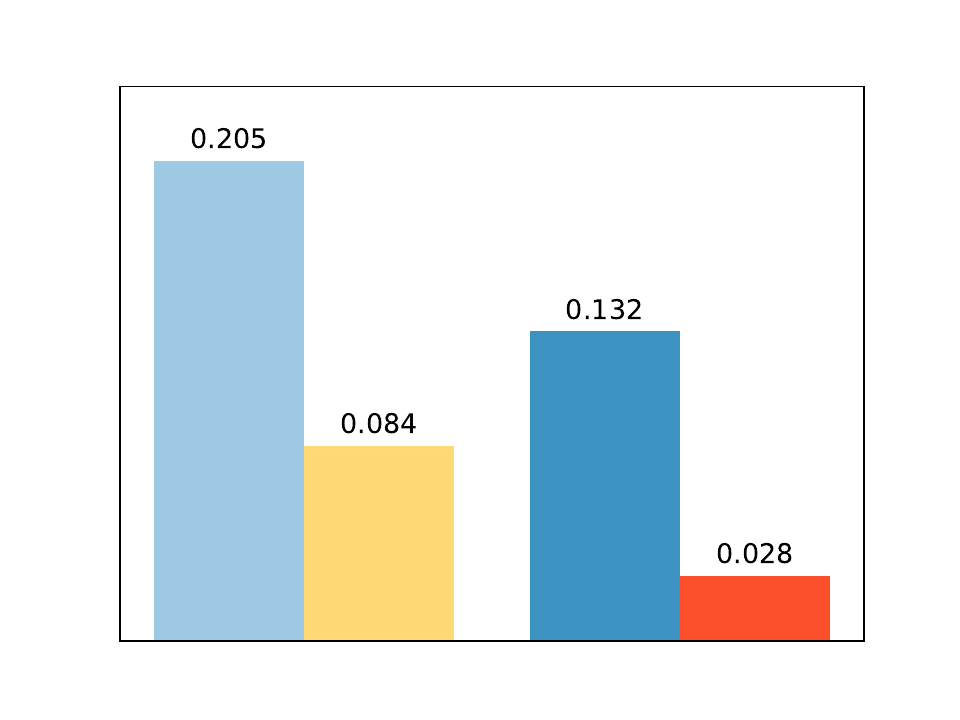}\\
\includegraphics[width=0.28\columnwidth,trim=0cm 1.5cm 3.5cm 0.5cm]{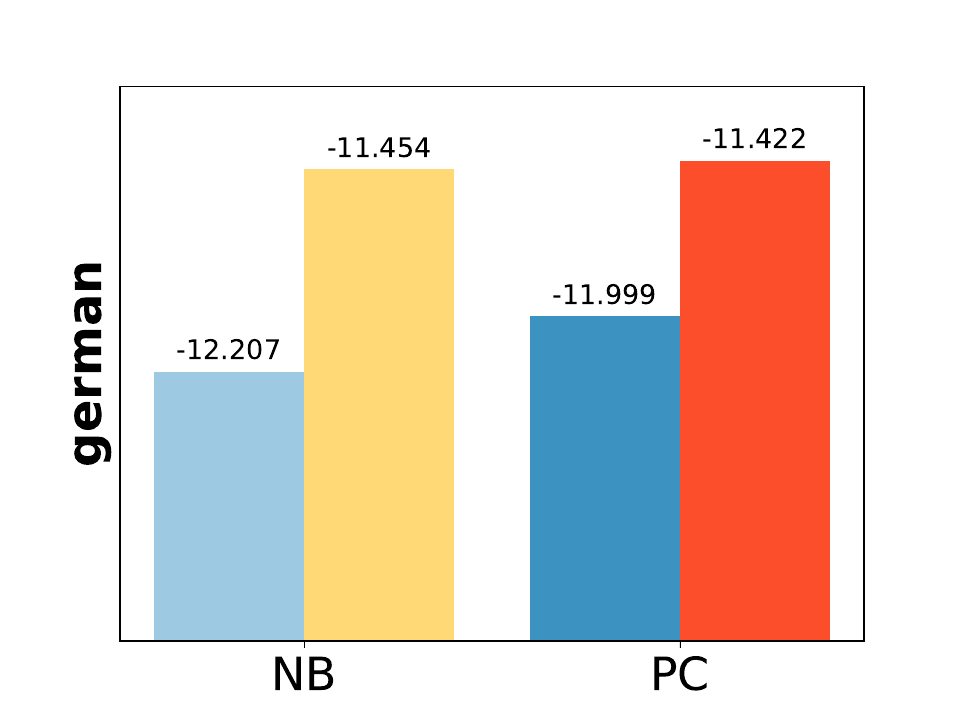}&
\includegraphics[width=0.28\columnwidth,trim=0cm 1.5cm 3.5cm 0.5cm]{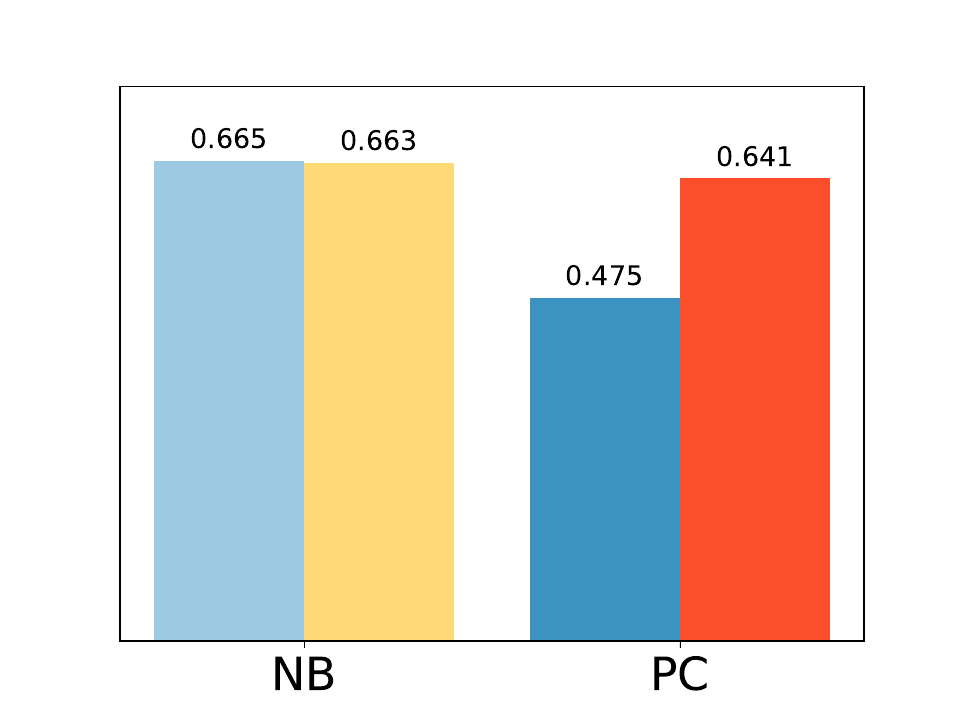}&
\includegraphics[width=0.28\columnwidth,trim=0cm 1.5cm 3.5cm 0.5cm]{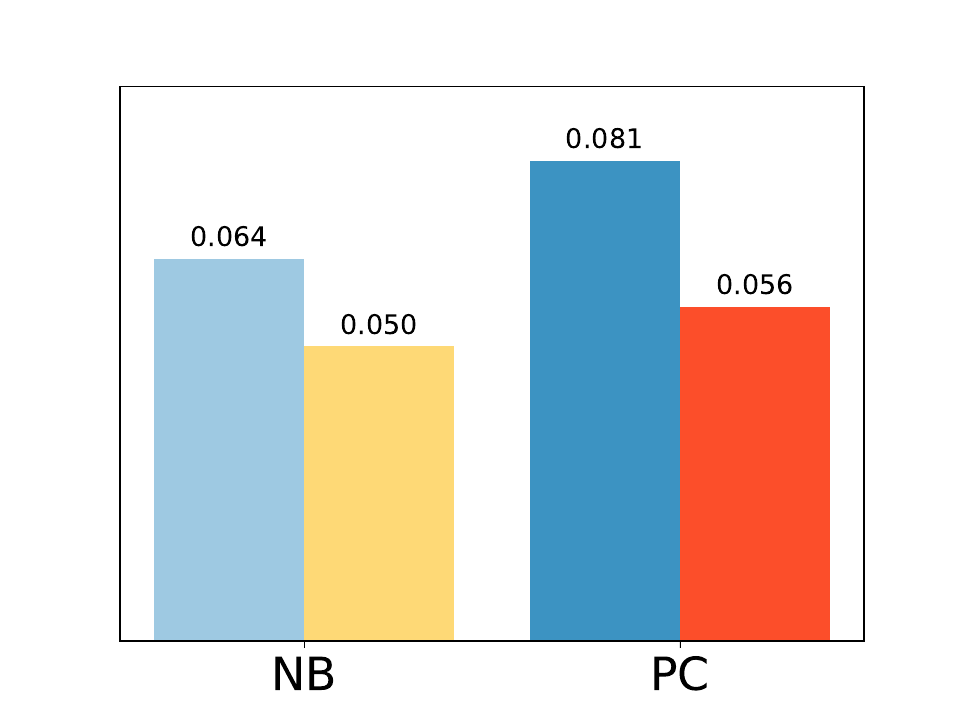}\\
\end{tabular}
\caption{Comparison of fair probability distributions. \textbf{Columns:} log-likelihood, F1-score, discrimination score (higher is better for the first two; lower is better for last). \textbf{Rows:} COMPAS, Adult, German datasets. The four bars in each graph from left to right are: 1) \twonb, 2) \latentnb, 3) \nlatent, 4) \fairpc.
\label{fig:exp-realworld-4base}}
\end{figure}

\subsection{Real-World Data}
\label{sec:exp-realworld}
\paragraph{Data}
We use three datasets: COMPAS~\cite{compas}, Adult, and German~\cite{Dua2019}, which are commonly studied benchmarks for fair ML.
They contain both numerical and categorical features and are used for predicting recidivism, income level, and credit risk, respectively.
We wish to make predictions that are fair with respect to a protected attribute: ``sex'' for Adult and German, and ``ethnicity'' for COMPAS.
As pre-processing, we discretize numerical features (e.g.\ age), remove unique or duplicate features (e.g.\ names of individuals), and remove low frequency counts.

\paragraph{Probabilistic methods}
We first compare against probabilistic methods to illustrate the effects of using latent variables and learning more expressive distributions.
Figure~\ref{fig:exp-realworld-4base} summarizes the result. The bars, from left to right, correspond to \twonb, \latentnb, \nlatent, and \fairpc. First and last two bars in each graph correspond to NB and PC models, respectively. Blue bars denote non-latent model, and yellow/orange denote latent-variable approach.

In terms of log-likelihoods, both PC-based methods outperform NB models, which aligns with our motivation for relaxing the naive Bayes assumption---to better fit the data distribution.
%
Furthermore, models with latent variables outperform their corresponding non-latent models, i.e., \latentnb\ outperforms \twonb\ and \fairpc\ outperforms \nlatent. This validates our argument made in Section~\ref{sec:model} that the latent variable approach can achieve higher likelihood than enforcing fairness directly in the observed label.
Next, we compare the methods using F1-score as there is class imbalance in these datasets. Although it is measured with respect to possibly biased labels, \fairpc\ achieves competitive performance, demonstrating that the latent fair decision variable still exhibits high similarity with the observed labels.
Lastly, \fairpc\ achieves the lowest discrimination scores in COMPAS and Adult datasets by a significant margin. As expected, PCs also achieve lower discrimination scores than their counterpart NB models, as they fit the data distribution better.

\begin{figure}
    \centering
    \begin{tabular}{ll}
    \includegraphics[width=0.49\columnwidth]{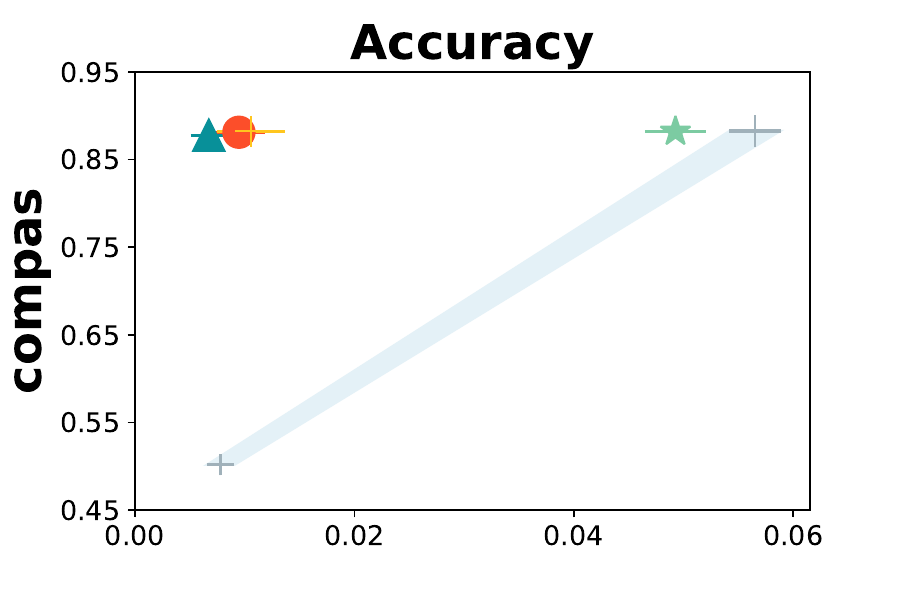}&
    \includegraphics[width=0.49\columnwidth]{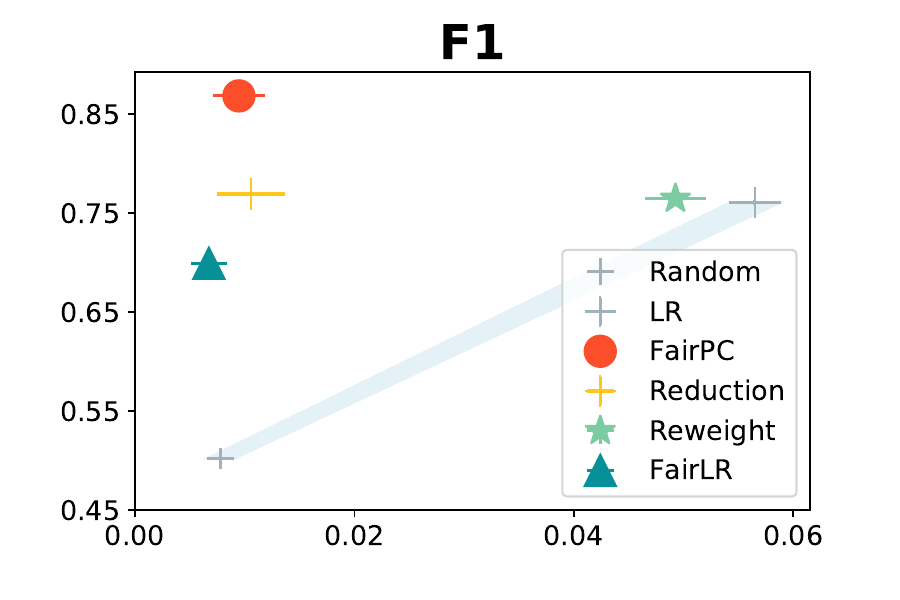}\\
    \includegraphics[width=0.49\columnwidth]{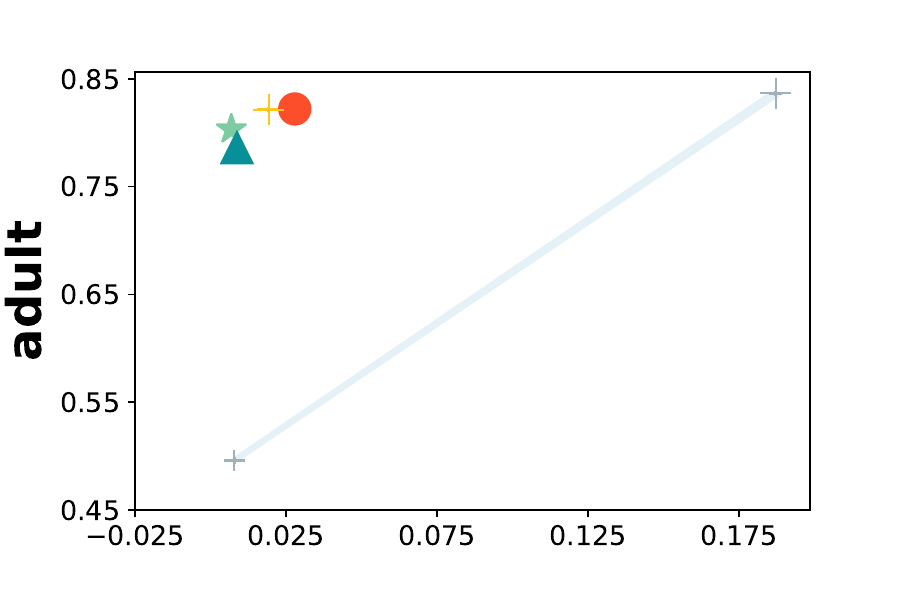}&
    \includegraphics[width=0.49\columnwidth]{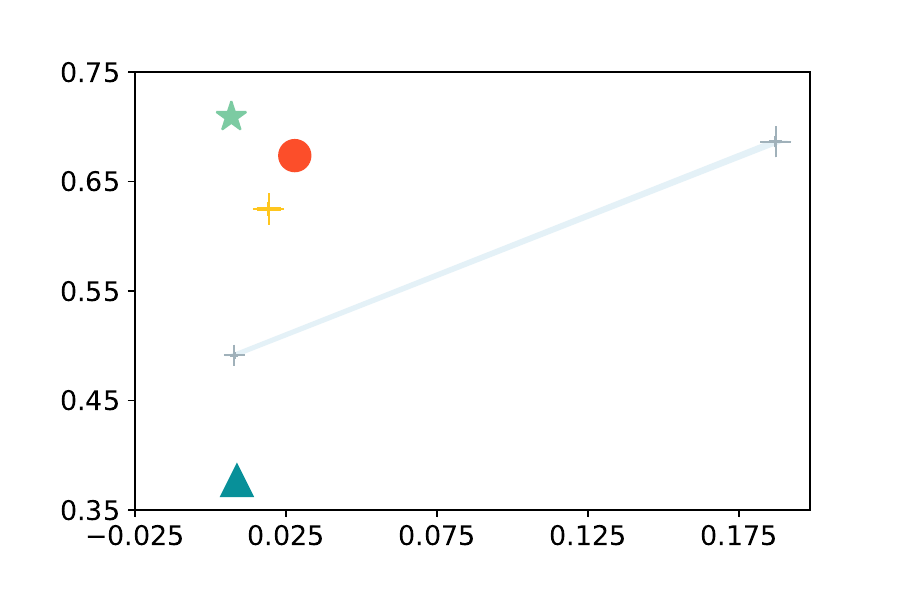}\\
    \includegraphics[width=0.49\columnwidth]{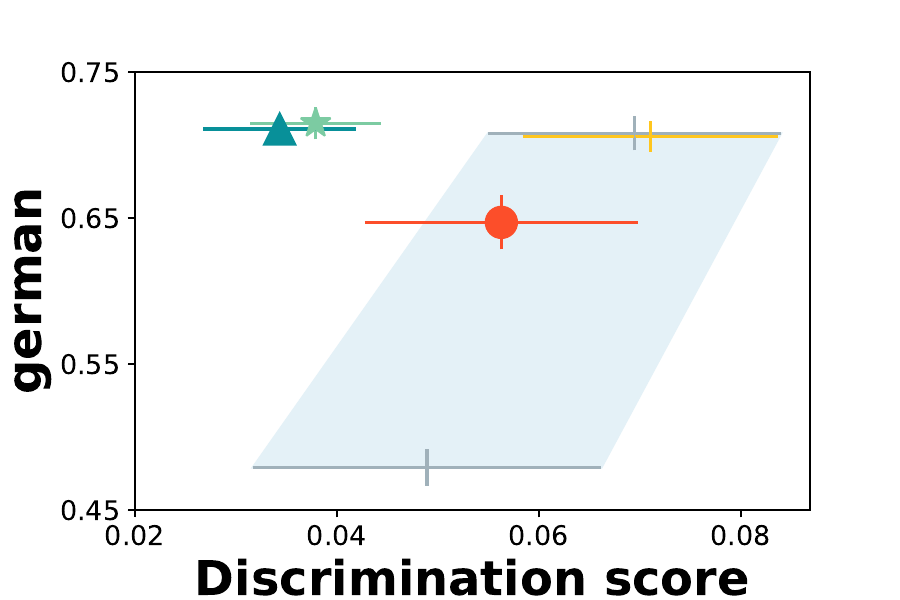}&
    \includegraphics[width=0.49\columnwidth]{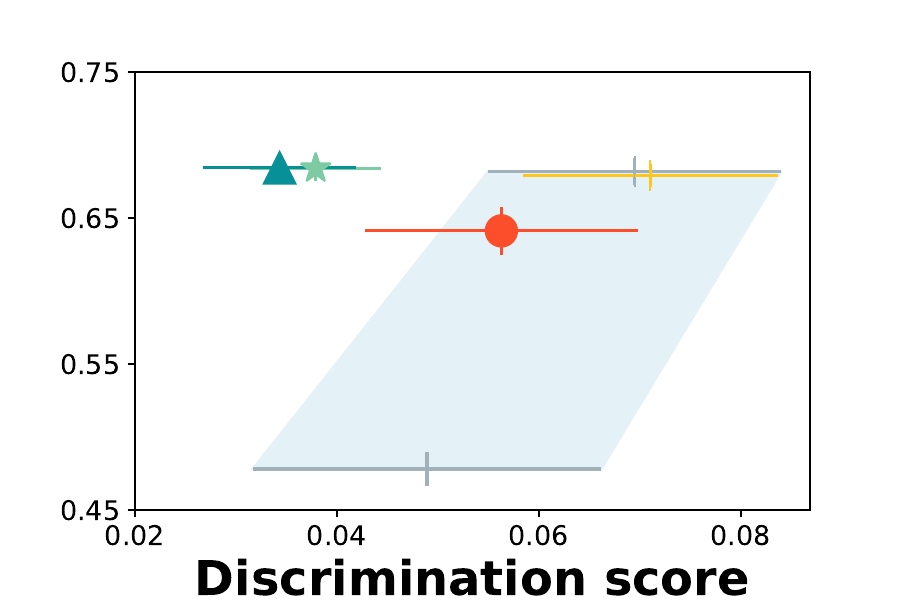}
    \end{tabular}
    \caption{Predictive performance (y-axis) vs.\ discrimination score (x-axis) for \fairpc\ and fair classification methods (\fairclass, \fairreduct, \reweight), in addition with two trivial baselines (\rand\ and \lr). \textbf{Columns:} accuracy, F1-score.
    \textbf{Rows:} COMPAS, Adult, German datasets. \label{fig:exp-rw-benchmark}}
\end{figure}

\paragraph{Discriminative classifiers}
Next we compare \fairpc\ to existing fair classification methods. Figure~\ref{fig:exp-rw-benchmark} shows the trade-off between predictive performance and fairness. We add two other baselines to the plot: \rand, which makes random predictions, and \lr, which is an unconstrained logistic regression classifier. They represent the two ends of the fairness-accuracy tradeoff. \rand\ has no predictive power but low discrimination, while \lr\ has high accuracy but unfair. Informally, the further above the line between these baselines, the better the method optimizes this tradeoff.

On COMPAS and Adult datasets, our approach achieves a good balance between predictive performance and fairness guarantees. In fact, it achieves the best or close to best accuracy and F1-score, again showing that the latent decision variable is highly similar to the observed labels even though the explicit objective is not to predict the unfair labels. 
However, on German dataset, while \fairclass\ and \reweight\ achieve the best performance on average, the estimates for all models including the trivial baselines are too noisy to draw a statistically significant conclusion. This may be explained by the fact that the dataset is relatively small with 1000 samples.

\subsection{Synthetic Data}
\label{sec:exp-synthetic}
As discussed previously, ideally we want to evaluate against the true target labels, but they are generally unknown in real-world data. Therefore, we also evaluate on synthetic data with fair ground-truth labels in order to evaluate whether our model indeed captures the hidden process of bias and makes accurate predictions.

\paragraph{Generating Data}
We generate data by constructing a fair PC $\PC_{true}$ to represent the ``true distribution'' and sampling from it.
The process that generates biased labels $d$ is represented by the following (conditional) probability table:
\begin{center}
\scalebox{0.76}{
    \footnotesize
    \begin{tabular}{c|cc||c|cccc}
    \toprule
    $\cdot$ & $D_f$ & $S$ & $d_f, s$ & 1,1 & 1,0 & 0,1 &0,0\\
    \midrule
    \normalsize
    $\Pr(\cdot\!=\!1)$ & 0.5 & 0.3 & $\Pr(D\!=\!1\mid D_f\!=\!d_f,S\!=\!s)$& 0.8 & 0.9 & 0.1 & 0.4 \\
    \bottomrule
\end{tabular}
}
\end{center}
Here, $S=1$ is the minority group, and the unfair label $D$ is in favor of the majority group: $D$ is more likely to be positive for the majority group $S\!=\!0$ than for $S\!=\!1$, for both values of fair label $D_f$ but at different rates.
To evaluate on a wide range of datasets, we randomly generate the sub-circuits of $\PC_{true}$ over features $\Xs$ as tree distributions, randomly initializing the parameters with Laplace smoothing.
We generated different synthetic datasets with the number of non-sensitive features ranging from 10 to 30, using 10-fold CV for each.

\begin{figure}[t]
  \centering
\begin{tabular}{cc}
  \includegraphics[width=0.45\columnwidth]{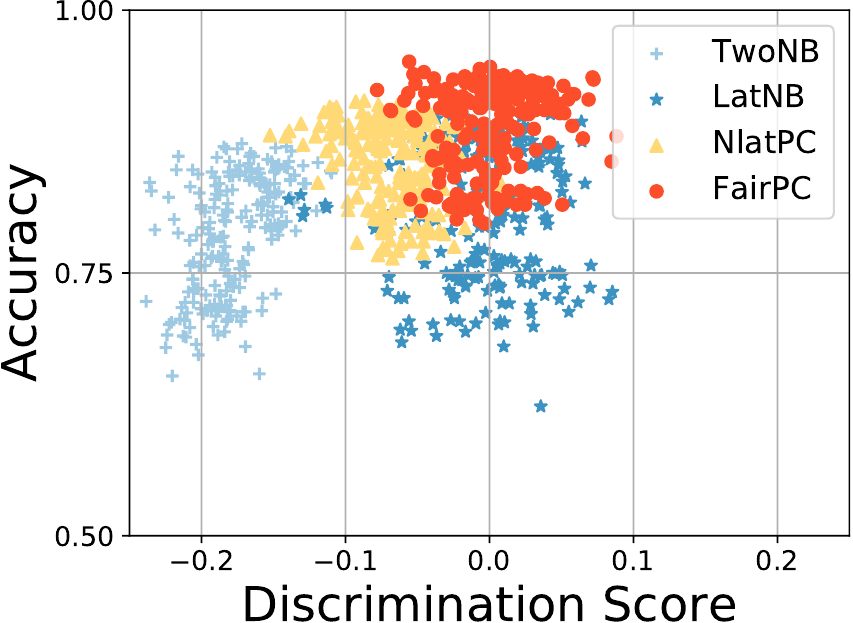}&
  \includegraphics[width=0.45\columnwidth]{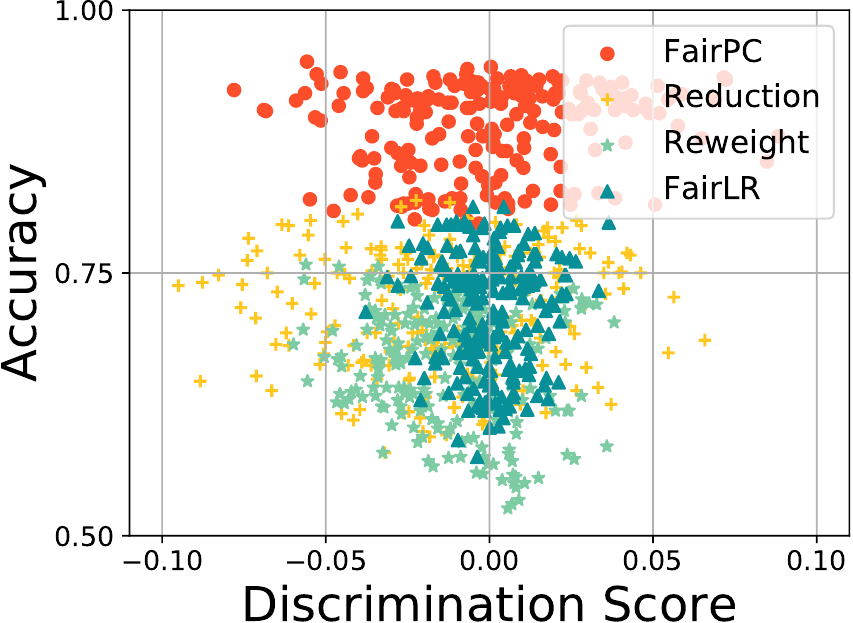}\\
\end{tabular}
  \captionof{figure}{\label{fig:exp-sync}Accuracy (y-axis) vs.\ discrimination score (x-axis) on synthetic datasets. We compare \fairpc\ with \twonb, \latentnb, \nlatent\ (left) and with \fairreduct, \reweight, \fairclass\ (right). Each dot is a single run on a generated dataset using the method indicated by its color.}
\end{figure}

\paragraph{Results}

We first test \fairpc, \latentnb, \nlatent\ and \nlatent\ on the generated datasets. Figure~\ref{fig:exp-sync} (left) illustrates the accuracy and discrimination scores on separate test sets with fair decision labels.

In terms of accuracy, PCs outperform NBs, and latent variable approaches outperform non-latent ones, which shows that adopting density estimation to fit the data and introducing a latent variable indeed help improve the performance.

When comparing the average discrimination score for each method, \twonb\ and \nlatent\ always have negative scores, showing that the non-latent methods are more biased towards the majority group; while \latentnb\ and \fairpc\ are more equally distributed around zero on the x-axis, thus demonstrating that a latent fair decision variable helps to correct this bias. While both latent variable approaches achieve reasonably low discrimination on average, \fairpc\ is more stable and has even lower average discrimination score than \latentnb.
Moreover \fairpc\ also outperforms the other probabilistic methods in terms of likelihood; 
see Appendix~\ref{sec:app-exp}.

We also compare \fairpc\ to \fairclass, \fairreduct, and \reweight, the results visualized in Figure~\ref{fig:exp-sync} (right). Our method achieves a much higher accuracy w.r.t.\ the generated fair labels; for instance, the average accuracy of \fairpc\ is around 0.17 higher than that of \fairclass.
Also, we are still being comparable in terms of discrimination score, illustrating the benefits of explicitly modeling the latent fair decision.

\subsection{Additional experiments}
Appendix~\ref{sec:app-exp} includes learning curves, statistical tests, and detailed performance of our real-world data experiments, as well as the following additional experiments.
We empirically validated that initializing parameters using prior knowledge as described in Section~\ref{sec:param-learn} indeed converges closer to the true distribution of $\Pr(D \given S,D_f)$ than randomly initializing parameters.
In addition, as mentioned in Section~\ref{sec:param-learn}, our method can be applied even on datasets with missing values, with no change to the algorithm. We demonstrate this empirically and show that our approach still gets comparably good performance for density estimation.

\section{Conclusion}
In this paper, we proposed a latent variable approach to learning fair distributions that satisfy demographic parity, and developed an algorithm to learn fair probabilistic circuits from incomplete data.
Experimental evaluation on simulated data showed that our method consistently achieves the highest log-likelihoods and a low discrimination score. It also accurately predicts true fair decisions, and even on real-world data where fair labels are not available, our predictions remain close to the unfair ones.

\subsubsection*{Acknowledgments}
This work is partially supported by NSF grants \#IIS-1943641, \#IIS-1633857,
\#CCF-1837129, DARPA grant \#N66001-17-2-4032, a Sloan Fellowship, Intel, and Facebook.

\bibliography{references}

\clearpage
\appendix
\section{Parameter Learning using Expected Flows}
\label{sec:app-EF}
Here we formally define the circuit flow and expected flows, and provide details on EM parameter learning using expected flows as well as proof of correctness. We assume smooth, decomposable, and deterministic PCs in the following sections.

\subsection{Definitions}

\begin{defn}[Context]\label{def:ctx}
    Let $\PC$ be a PC over RVs $\Zs$ and $n$ be one of its nodes. The \emph{context} $\gamma_n$ of node $n$ denotes all joint assignments that return a nonzero value for all nodes in a path between the root of $\PC$ and $n$. 
    \begin{equation*}
        \gamma_n := \bigcup_{p\in\pa(n)} \gamma_p \cap \supp(n)
    \end{equation*}
    where $\pa(n)$ refers to the parent nodes of $n$ and $\supp(n) := \{\zs : \PC_n(\zs)>0\}$ is the support of node $n$.
\end{defn}
Note that the context of a node is different from its support. Even if the node returns a non-zero value for some input, its output may be multiplied by 0 at its ancestor nodes; i.e., such node does not contribute to the circuit output of that assignment. 

We can now express circuit flows and expected flows in terms of contexts.
Intuitively, the context of a circuit node is the set of all complete inputs that ``activate'' the node. Hence, an edge is ``activated'' by an input if it is in the contexts of both nodes for that edge.
%
\begin{defn}[Circuit flow]
    Let $\PC$ be a PC over variables $\Zs$, $(n,c)$ its edge, and $\zs$ a joint assignment to $\Zs$. The \emph{circuit flow} of $(n,c)$ given $\zs$ is
    \begin{equation}
        F_{\zs}(n,c) = [\zs \in \gamma_n \cap \gamma_c]. \label{eq:flow-context}
    \end{equation}
\end{defn}

\begin{defn}[Expected flow]
    Let $\PC$ be a PC over variables $\Zs$, $(n,c)$ its edge, and $\es$ a partial assignment to $\Es \subseteq \Zs$. The \emph{expected flow} of $(n,c)$ given $\es$ is given by
    \begin{align}
        \EF_{\es,\theta}(n,c) := \Ex_{\zs \sim \Pr_\PC(\cdot \given \es)} [F_{\zs}(n,c)]. \label{eq:exp-flow}
    \end{align}
\end{defn}

Then the flow given a dataset is simply the sum of flows given each data point. That is, given a complete data $\D$, the circuit flow of $(n,c)$ is
\begin{align*}
    F_{\D}(n,c) = \sum_{\D_i \in \D} F_{\D_i}(n,c),
\end{align*}
where $\D_i$ is the i-th data point of $\D$, which must be a complete assignment $\zs$. Similarly, given an incomplete data $\D$, the expected flow of $(n,c)$ w.r.t.\ parameters $\theta$ is
\begin{align*}
    \EF_{\D,\theta}(n,c) = \sum_{\D_i \in \D} \EF_{\D_i,\theta}(n,c),
\end{align*}
where $\D_i$ may be a partial assignment $\es$ for some $\Es \subseteq \Zs$.

\subsection{Computing the Expected Flow}

Next we describe how to compute the expected flows. First, focusing on the expected flow given a single partial assignment $\es$, we can express the expected flow as the following using Equations~\ref{eq:flow-context} and \ref{eq:exp-flow}.
%
\begin{align}
    \EF_{\es,\theta}(n,c) = \Ex_{\zs \sim \Pr_\PC(\cdot \given \es)} [\zs \in \gamma_n \cap \gamma_c] = \pr_\PC(\gamma_n \cap \gamma_c \given \es)  \label{eq:ef-edge}
\end{align}
Furthermore, with determinism, the sub-circuit formed by ``activated'' edges for any complete input forms a tree~\cite{ChoiDarwiche17}. Thus, for a complete evidence $\zs$, a node $n$ has exactly one parent $p$ such that $F_\zs(p,n)=1$, or equivalently, $\zs \in \gamma_p \cap \gamma_n$. Thus,
\begin{align}
    \sum_{p\in\pa(n)} \EF_{\es,\paras}(p,n)
    &= \sum_{p\in\pa(n)} \pr_\PC(\gamma_p \cap \gamma_n \given \es) \nonumber \\
    &= \pr_{\PC}(\gamma_n \given \es). \label{eq:ef-node}
\end{align}
We can observe from Equations~\ref{eq:ef-edge} and \ref{eq:ef-node} that if $\sum_{p\in\pa(n)} \EF_{\es,\paras}(p,n)=0$, then we also have $\EF_{\es,\paras}(n,c)=0$.

For an edge $(n,c)$ where $n$ is a sum node,
\begin{align}
    \EF_{\es,\paras}(n,c)
    &= \pr_{\PC}(\gamma_n \cap \gamma_c \given \es)
    = \frac{\pr_{\PC}(\es , \gamma_c \given \gamma_n) \pr_{\PC}(\gamma_n)}{\pr_{\PC}(\es)} \nonumber\\
    &= \frac{\pr_{\PC}(\gamma_n \given \es) \pr_{\PC}(\es , \gamma_c \given \gamma_n) \pr_{\PC}(\gamma_n) }{\pr_{\PC}(\gamma_n \given \es) \pr_{\PC}(\es)} \nonumber\\
    &= \frac{\pr_{\PC}(\gamma_n \given \es) \pr_{\PC}(\es , \gamma_c \given \gamma_n)}{\pr_{\PC}(\es \given \gamma_n)} \nonumber\\
    &= \left(\sum_{p\in\pa(n)} \EF_{\es,\paras}(p,n)\right) \frac{\theta_{n,c} \pr_{c}(\es)}{\pr_{n}(\es)}.     \label{eq:ef-sum}
\end{align}
Here, $\pr_n$ and $\pr_c$ refer to the distribution defined by the sub-circuits rooted at nodes $n$ and $c$, respectively. Because $\es$ can be partial observations, these probability corresponds to marginal queries.
For a smooth and decomposable probabilistic circuit, the marginals given a partial input for all circuit nodes can be computed by a single bottom-up evaluation of the circuit~\cite{darwiche2002knowledge}.
This amounts to marginalizing the leaf nodes according to the partial input (i.e., plugging in 1 for unobserved variables) and evaluating the circuit according to its recursive definition.

For an edge $(n,c)$ where $n$ is a product node, we have $\gamma_n \subseteq \gamma_c$ as follows:
\begin{align}
    &\gamma_n = \gamma_n \cap \supp(n)
    \subseteq \gamma_n \cap \supp(c) \label{eq:ctx-supp}\\
    &\subseteq \bigcup_{p \in \pa(c)} \gamma_p \cap \supp(c) = \gamma_c \nonumber
\end{align}
where Equation~\ref{eq:ctx-supp} follows from Definition~\ref{def:ctx} and the fact that any assignment that leads to a non-zero output for $n$ must also output non-zero for $c$ (i.e.\ $\supp(n) \subseteq \supp(c)$).
Then we can write the expected flow of $(n,c)$ as the following:
\begin{align}
    \EF_{\es,\paras}(n,c)
    &= \pr_{\PC}(\gamma_n \cap \gamma_c \given \es)
    = \pr_{\PC}(\gamma_n \given \es) \nonumber \\
    &= \sum_{p\in\pa(n)} \EF_{\es,\paras}(p,n).  \label{eq:ef-product}
\end{align}
Therefore, Equations~\ref{eq:ef-sum} and \ref{eq:ef-product} describe how expected flow on edge $(n,c)$ can be computed using the expected flows from parents of $n$ and the marginal probabilities at nodes $n$ and $c$.
We can thus compute the the expected flow via a bottom-up evaluation (to compute the marginals) followed by a top-down pass as shown in Algorithm~\ref{alg:exp-flow}.
We cache intermediate results to avoid redundant computations and to ensure a linear-time evaluation.

\begin{algorithm}[tb]
\SetAlgoLined
    \SetKwInOut{Input}{Input}
    \SetKwInOut{Output}{Output}
    \Input{PSDD $\PC$, one data sample $d$, marginal likelihood $\pr_{\PC}$ cached from bottom-up pass}
    \Output{Expected flow of sample $d$ for each node and edge, cached in $\EF$}
    // traverse PSDD nodes by visiting parents before children\\
    \For{n in PSDD $\PC$}{
        \uIf{$n$ is root}{
            $\EF(n) \leftarrow 1$
        }\Else{
            $\EF(n) \leftarrow \sum_{p \in \pa(n)} \EF(p,n)$
        }
        \uIf{n is a sum node}{
            \For{$c$ in $\ch(n)$}{
                $\EF(n,c) \leftarrow \EF(n)
                \cdot \frac{\theta_{n,c} \cdot \pr_{c}(d)}{\pr_{n}(d)}$
            }
        }
        \ElseIf{n is a product node}{
            \For{$c$ in $\ch(n)$}{
                $\EF(n,c) \leftarrow \EF(n)$
            }
        }
    }
 \caption{Computing the expected flow}
 \label{alg:exp-flow}
\end{algorithm}

To compute the expected flow on a dataset, we can compute the expected flow of each data sample in parallel via vectorization, and then simply sum the results per edge.

\subsection{Proof of Proposition~\ref{prop:em}}
We now prove Proposition~\ref{prop:em} which states that the following parameter update rule using expected flows is equivalent to an iteration of EM parameter learning for smooth, decomposable, and deterministic probabilistic circuits; i.e., equivalent to completing the dataset with weights then computing the maximum-likelihood parameters.
\begin{equation*}
    \theta_{n,c}^{\text{(new)}} = \EF_{\D,\paras}(n,c) / \sum_{c\in\ch(n)} \EF_{\D,\paras}(n,c).
\end{equation*}
Completing a dataset $\D$ with missing values, given a distribution $\Pr_{\paras}(.)$, amounts to constructing an auxiliary dataset $\D^\prime$ as follows:
for each data sample $\D_i \in \D$, there are samples $\D_{i,k}^\prime \in \D^\prime$ for $k=1,\dots,m_i$ with weights $\alpha_{i,k}$ such that each $\D_{i,k}^\prime$ is a full assignment that agrees with $\D_i$. Moreover, the weights are defined by the given distribution as: $\alpha_{i,k} = \Pr_{\paras}(\D_{i,k}^\prime \given \D_i)$.
Then the max-likelihood parameters of a circuit given this completed dataset $\D^\prime$ can be computed as:
$$\theta_{n,c} = F_{\D^\prime}(n,c) / \sum_{c\in\ch(n)} F_{\D^\prime}(n,c).$$
Note that since $\D^\prime$ is an expected/weighted dataset, the flows $F_{\D^\prime}$ are real numbers as opposed to integers, which is the case when every sample has weight 1.
Specifically,
\begin{align*}
    F_{\D^\prime}(n,c)
    &= \sum_{\D_{i,k}^\prime \in \D^\prime} \alpha_{i,k} F_{\D_{i,k}^\prime}(n,c) \\
    &= \sum_{\D_i \in \D} \sum_{k=1}^{m_i} \pr_{\paras}(\D_{i,k}^\prime \given \D_i) F_{\D_{i,k}^\prime}(n,c) \\
    &= \sum_{\D_i \in \D} \sum_{\zs \models \D_i} \pr_{\paras}(\zs \given \D_i) F_{\zs}(n,c)
    = \EF_{\D,\paras}(n,c).
\end{align*}

%



\section{Additional Experiments}
\label{sec:app-exp}

\subsection{Real-world Data}
\label{sec:app-exp-realworld}

\paragraph{Detailed results} 
Table~\ref{tab:exp-realworld} reports the detailed results of experiments in Section~\ref{sec:exp-realworld}. It compares 7 methods in terms of (1) log-likelihood, (2) accuracy, (3) F1-score, and (4) discrimination score on real world datasets. Bold number indicates the best result among fair probability distributions: \fairpc, \latentnb, \nlatent\ and \twonb.
$\uparrow$ (resp.\ $\downarrow$) indicates that \fairpc\ achieves better (resp.\ worse) result than the corresponding fair classification method: \fairclass, \fairreduct\ or \reweight.

\paragraph{Statistical tests}
Table~\ref{tab:ll-pvalue} reports the pairwise Wilcoxon signed-rank test p-values for the comparisons of test log-likelihoods for each pair of probabilistic methods on real world datasets. Bold values indicate that two methods are statistically equivalent with confidence 99\%.

Table~\ref{tab:predict-pvalue} reports the pairwise McNemar's test p-values for the comparisons of test set prediction results for each pair of algorithms (columns) on all real world datasets (rows). Bold values indicate two methods make statistically equivalent predictions in terms of accuracy (similarity with the observed labels) with confidence 99\%.

\begin{figure*}[htb!]
  \centering
  \begin{tabular}{ccc}
\includegraphics[width=0.2\textwidth]{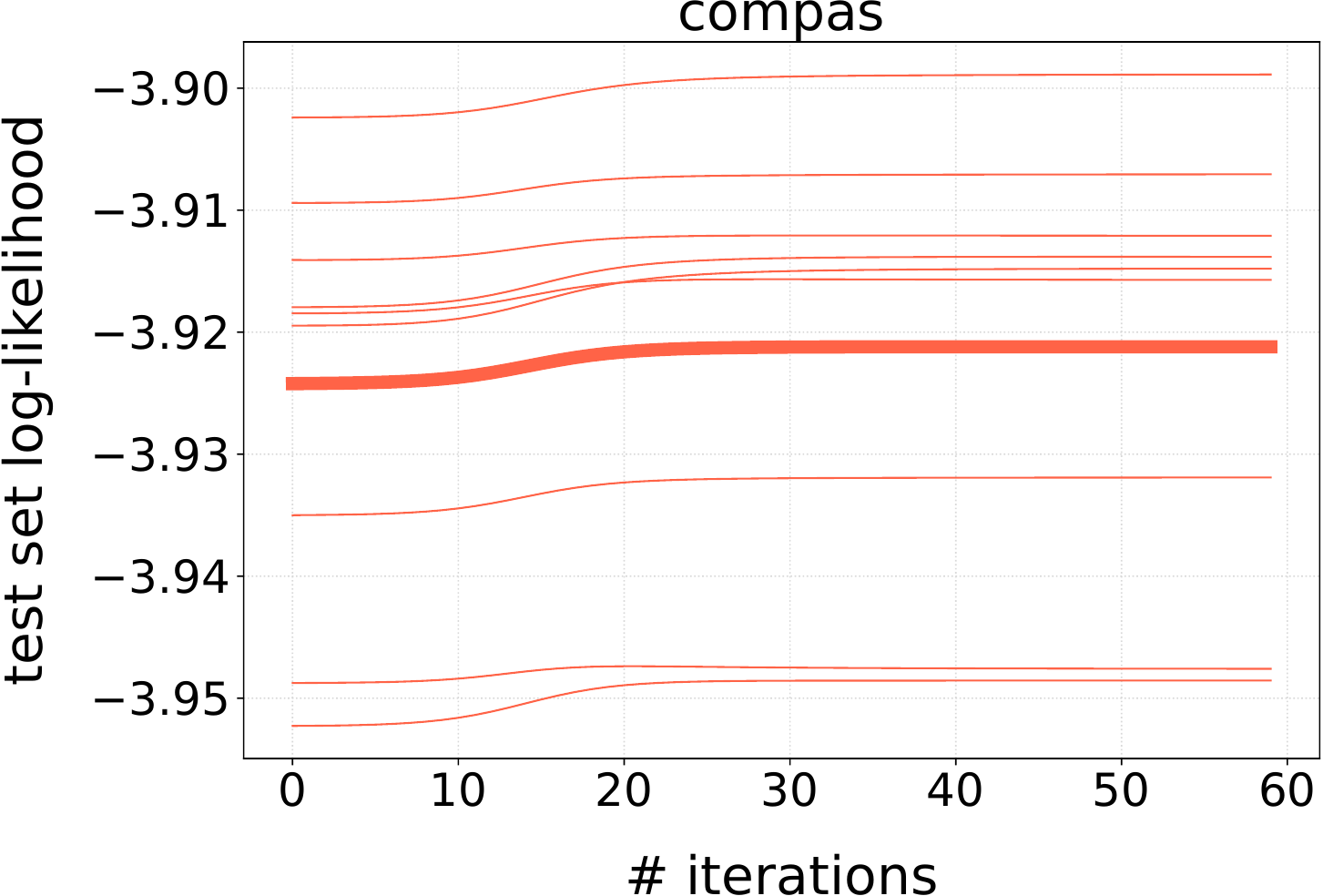}&
\includegraphics[width=0.2\textwidth]{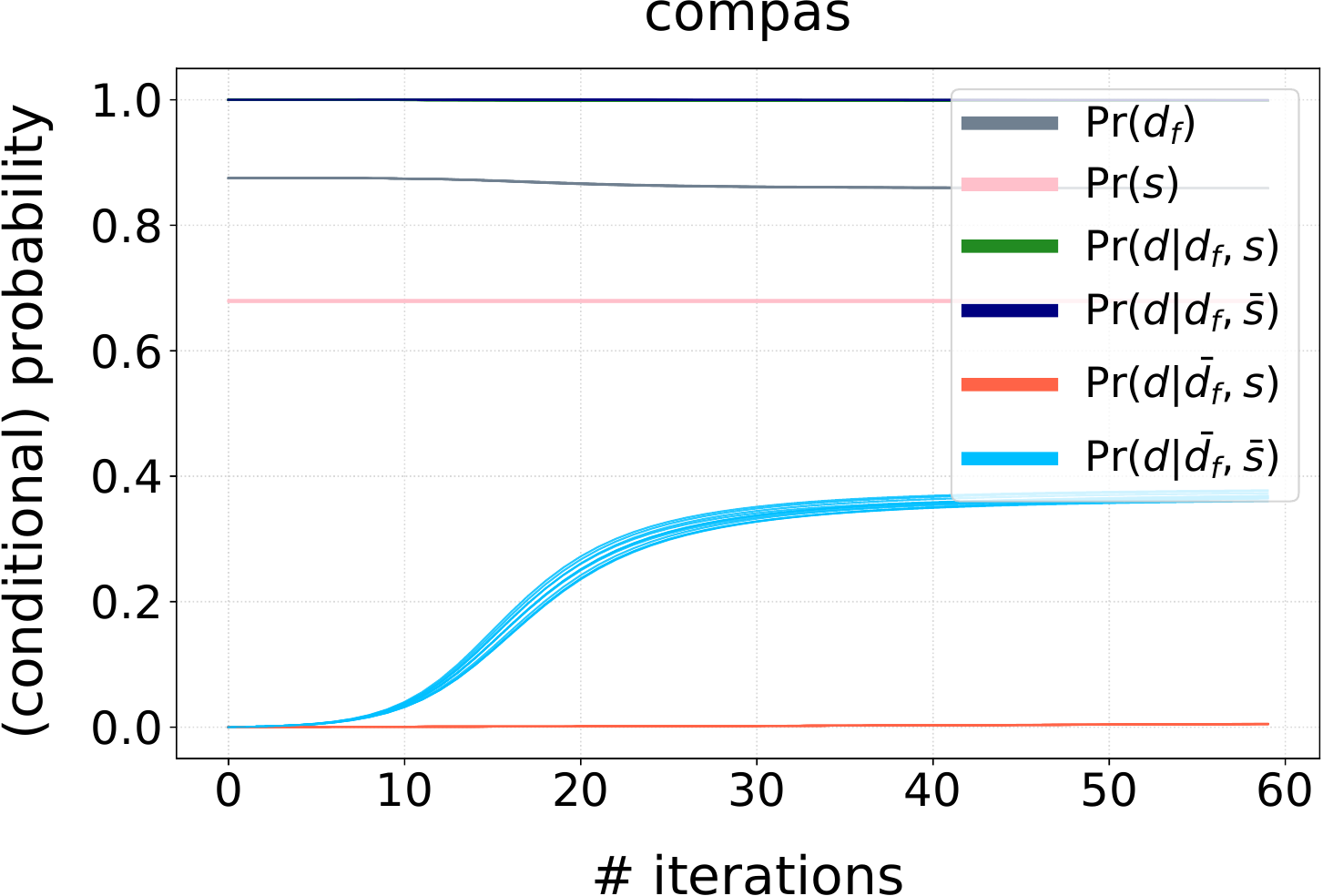}&  
\includegraphics[width=0.2\textwidth]{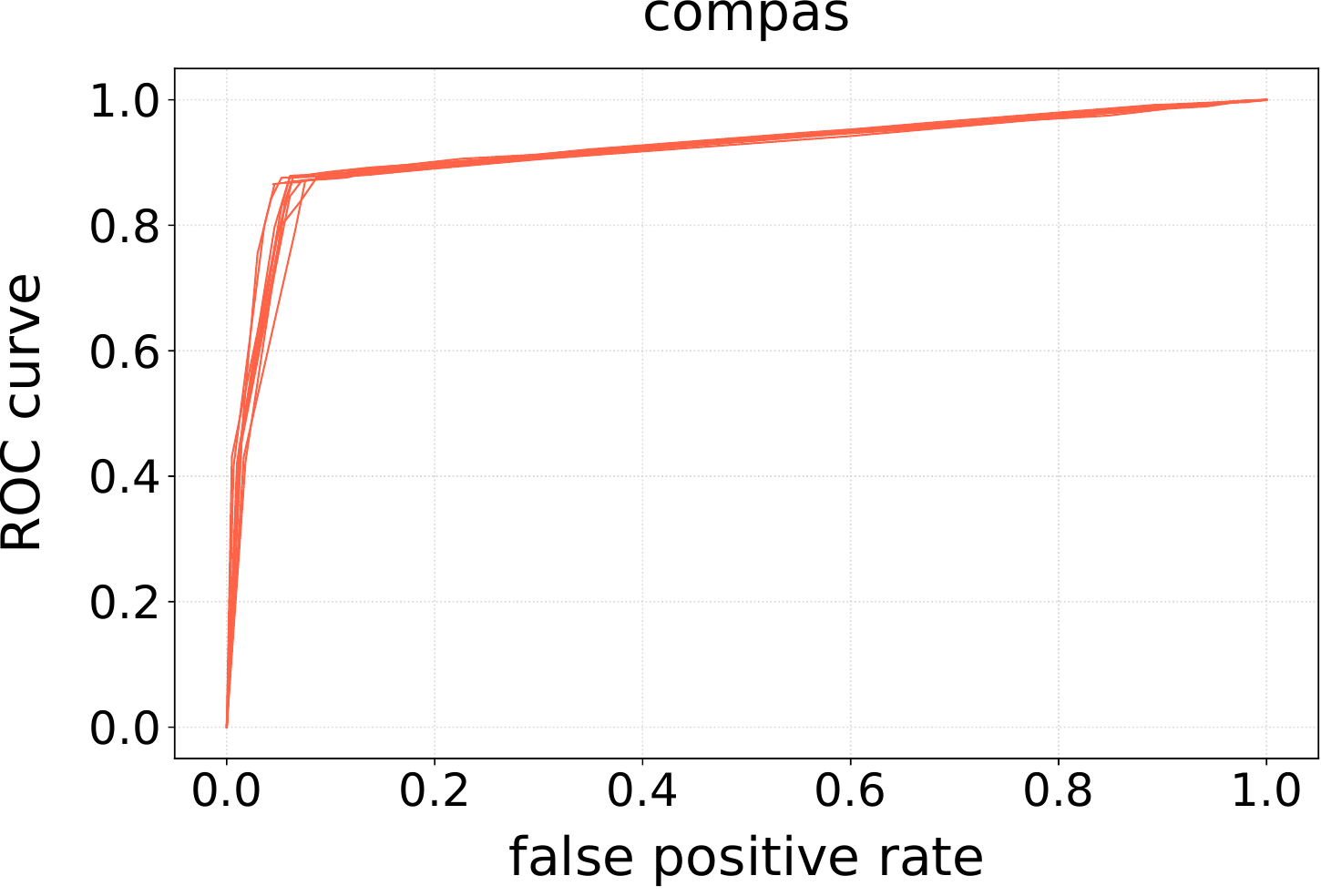}\\  
\includegraphics[width=0.2\textwidth]{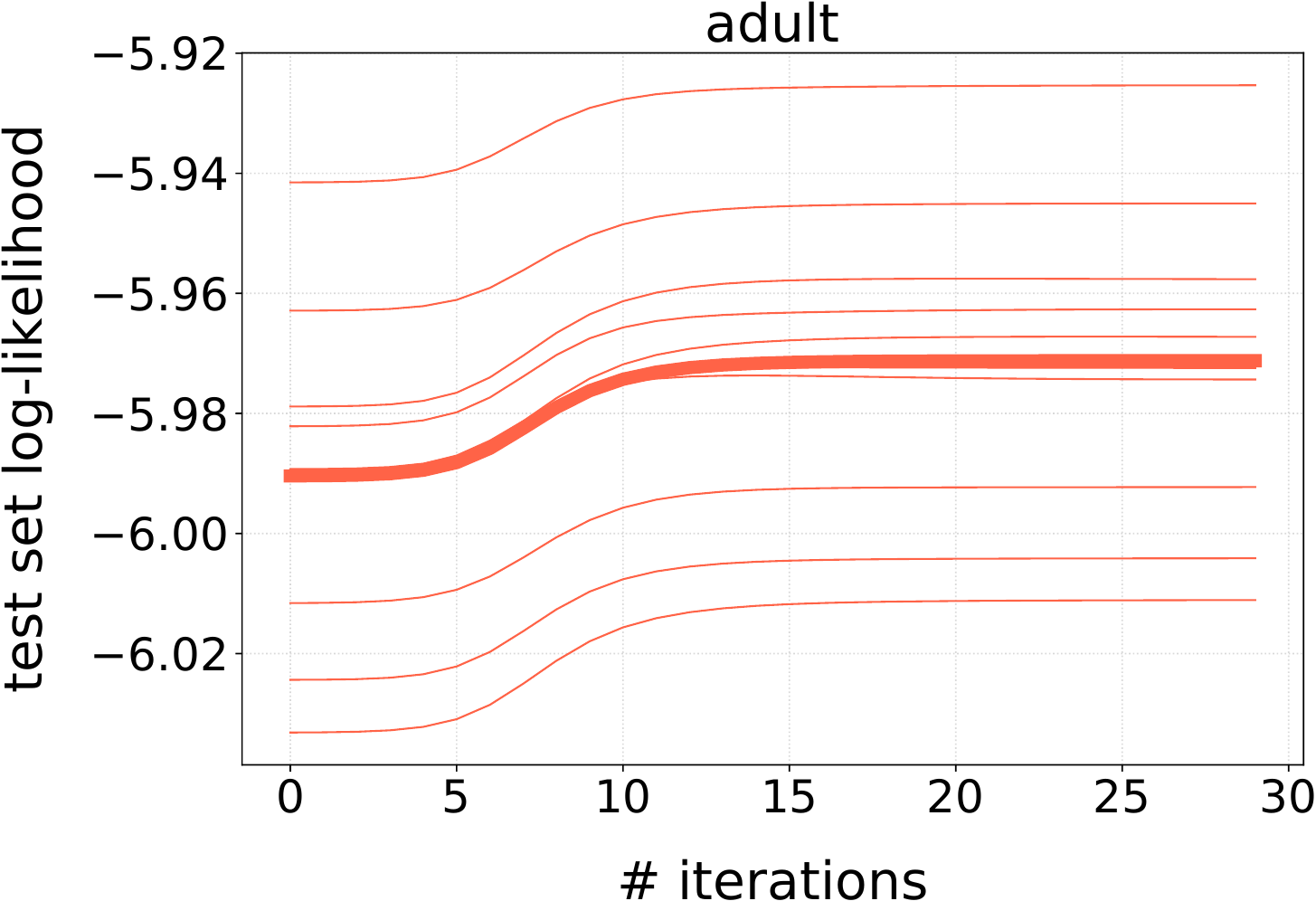}&  
\includegraphics[width=0.2\textwidth]{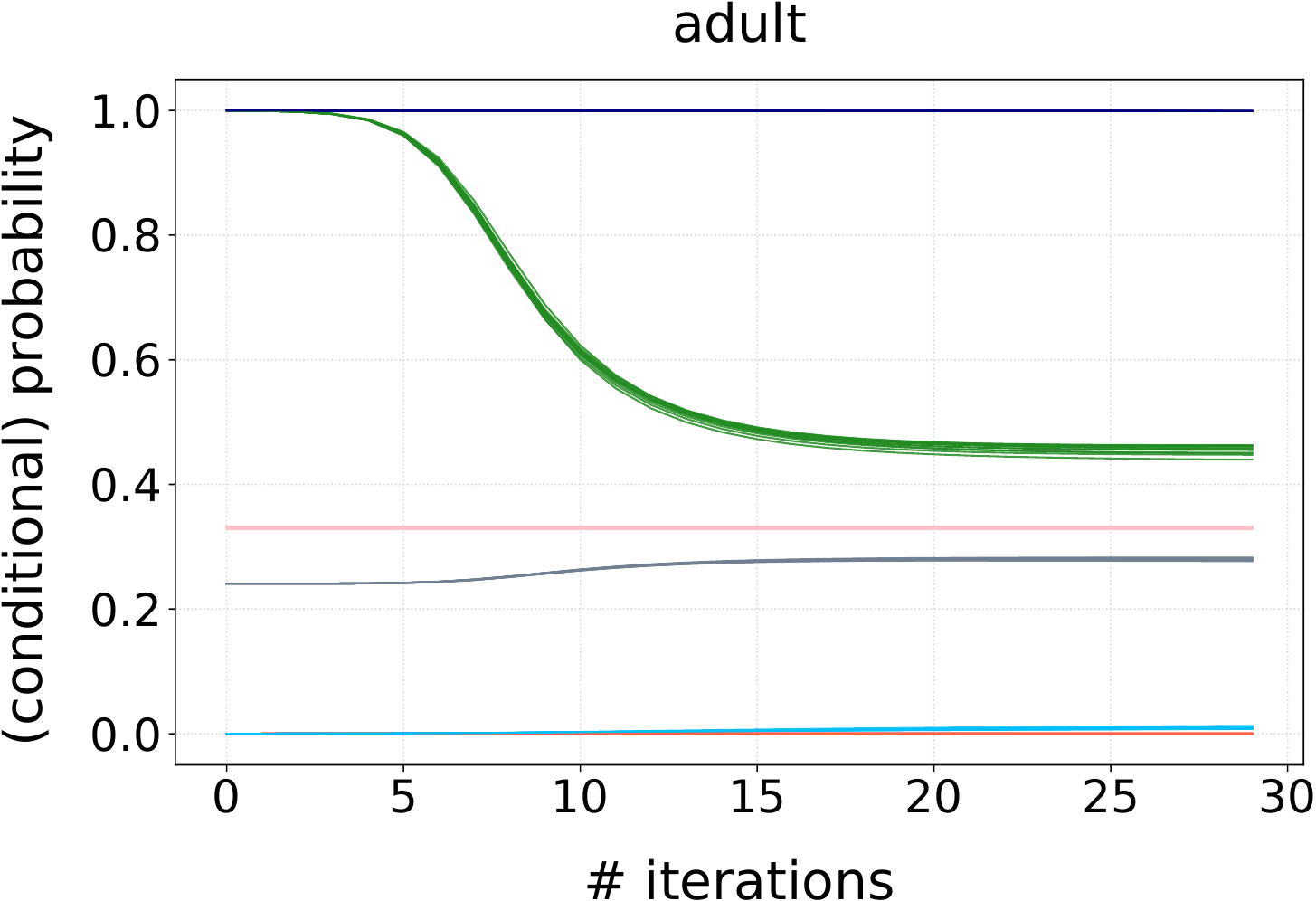}&  
\includegraphics[width=0.2\textwidth]{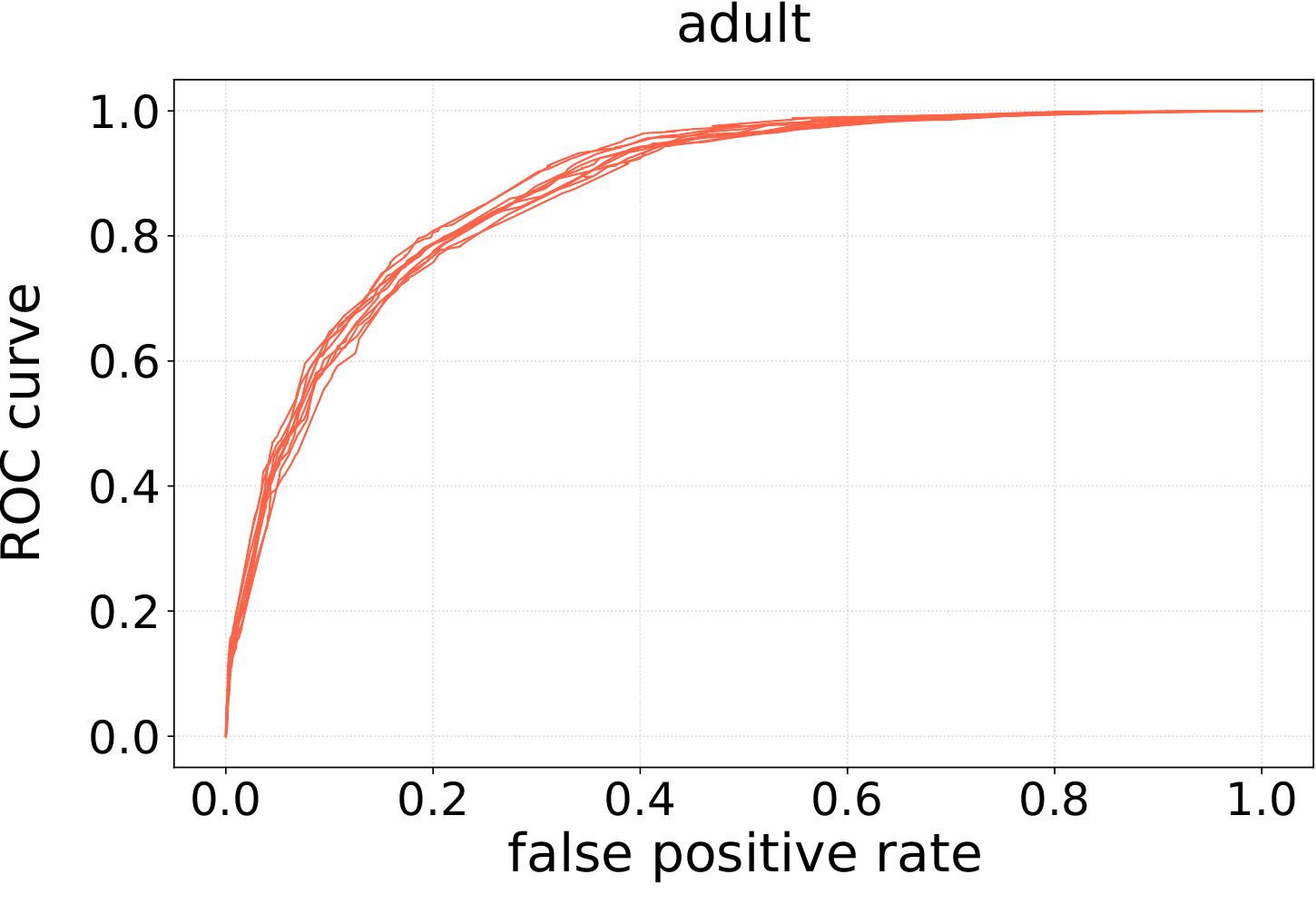}\\
\includegraphics[width=0.2\textwidth]{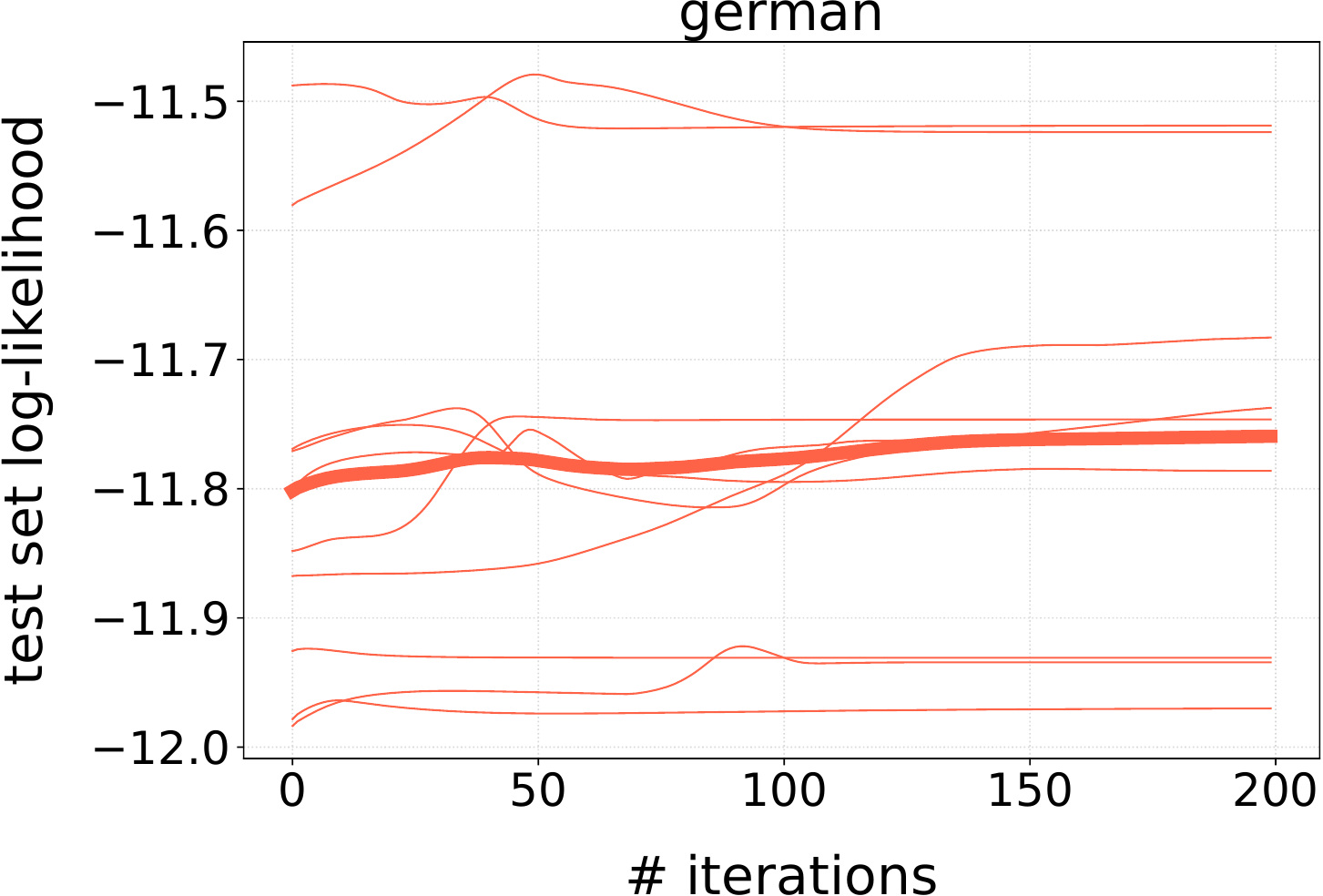}&  
\includegraphics[width=0.2\textwidth]{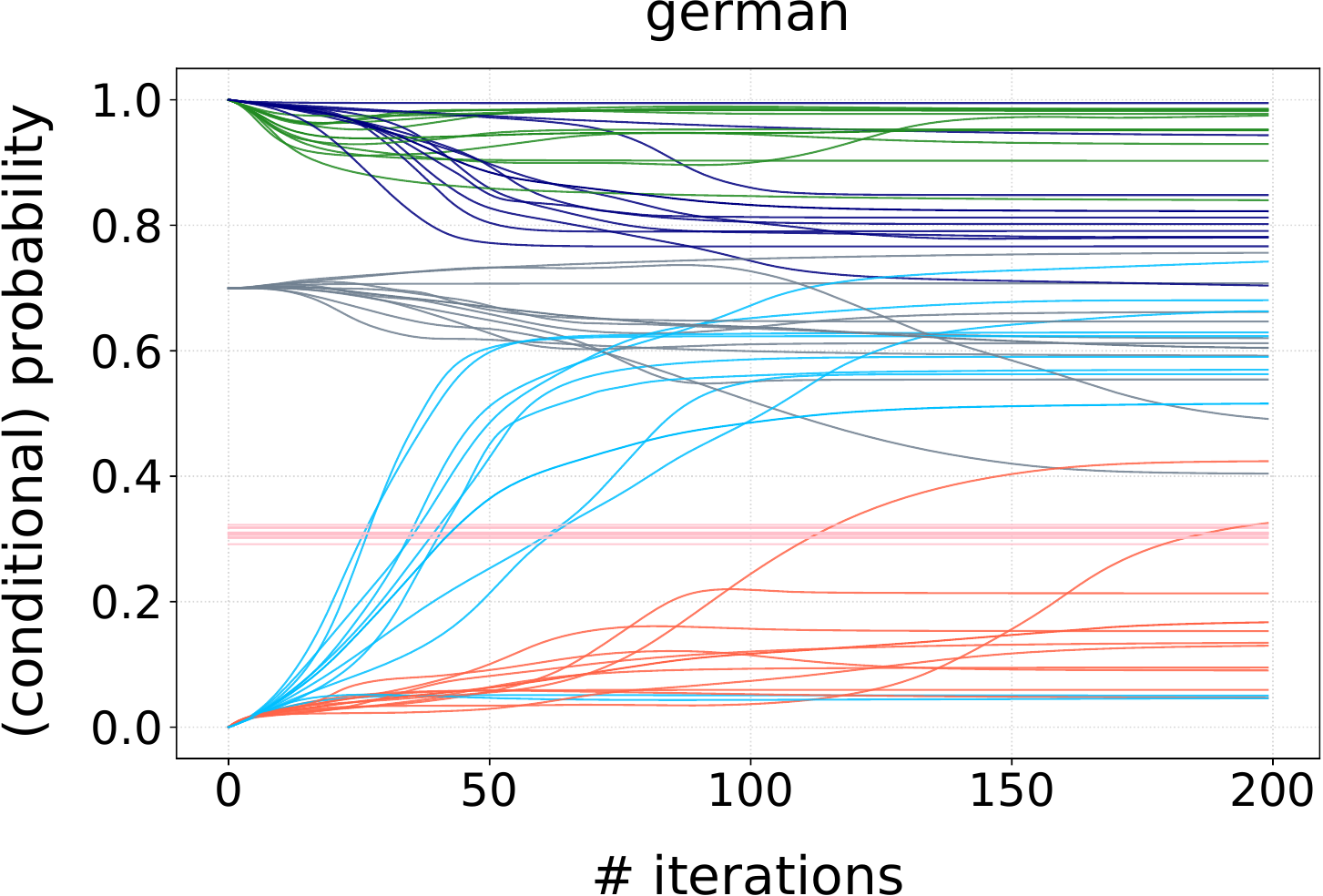}&  
\includegraphics[width=0.2\textwidth]{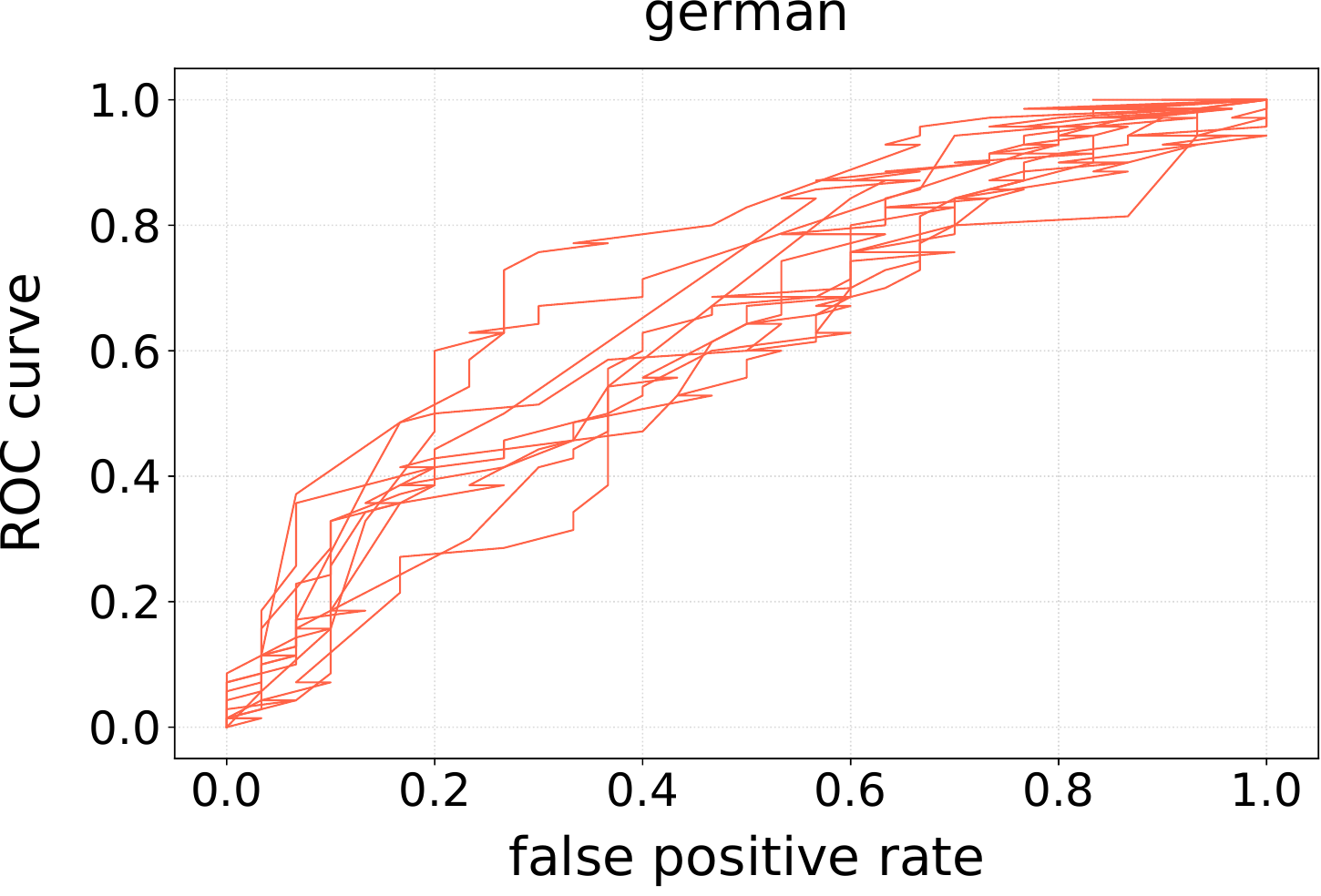}\\
\end{tabular}
  \caption{Training curves and ROC curves of \fairpc}
  \label{fig:fairpc-expall}
\end{figure*}

\paragraph{Learning curves}
Figure~\ref{fig:fairpc-expall} shows the 10-fold CV training curves (test log-likelihoods and probability table w.r.t number of iterations) and ROC curves of \fairpc, each line in the plot corresponding to one fold.
The test set log-likelihoods are reported after the structure is learned, and thus only describe the EM parameter learning iterations. 
Among the three datasets, German has the highest variance perhaps from having fewer number of examples.
On the other two datasets, we can observe that the learned parameters for the bias mechanism (i.e.\ for $D$, $D_f$, and $S$)
are fairly consistent across different CV folds.
For instance, the model learns that some negative labels ($D_f\!=\!0$) for the majority group ($S\!=\!0$) are flipped to positive labels in the observed data in COMPAS dataset, e.g., $P(D\!=\!1\given D_f\!=\!0, S\!=\!0)\!\approx\!0.4$; whereas in the case of Adult dataset, positive labels ($D_f\!=\!1$) of the minority group ($S\!=\!1$) are observed as negative labels with some probability, e.g., $P(D\!=\!0\given D_f\!=\!1, S\!=\!1)\!\approx\!0.6$.
Moreover, the ROC curves show the predictive performance on these datasets.

\subsection{Synthetic Data}
\label{sec:app-exp-syn}



\paragraph{Likelihood comparison}
Table~\ref{tab:synthetic-ll} reports the average test log-likelihoods w.r.t.\ different numbers of non-sensitive variables, comparing the true distribution (True), \latentnb, \nlatent, and \fairpc\ initialized from random start (\fairpc-rand) or prior knowledge (\fairpc-prior). This shows that \fairpc\ consistently achieves the best log-likelihoods, and \latentnb\ performs the worst. 

\begin{figure*}[!ht]
    \centering
    \begin{tabular}{cc}
    \includegraphics[width=0.5\columnwidth]{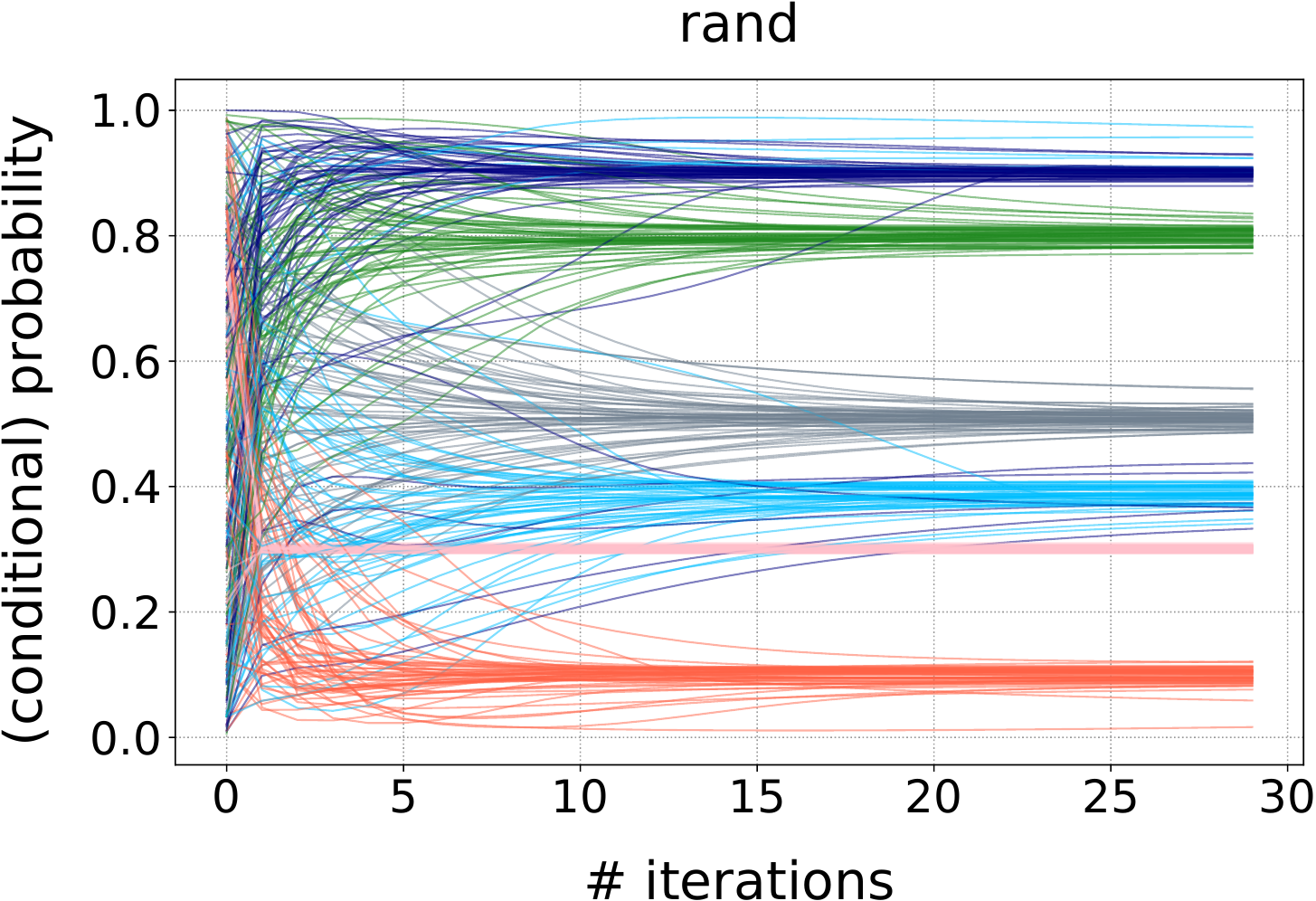}&
    \includegraphics[width=0.5\columnwidth]{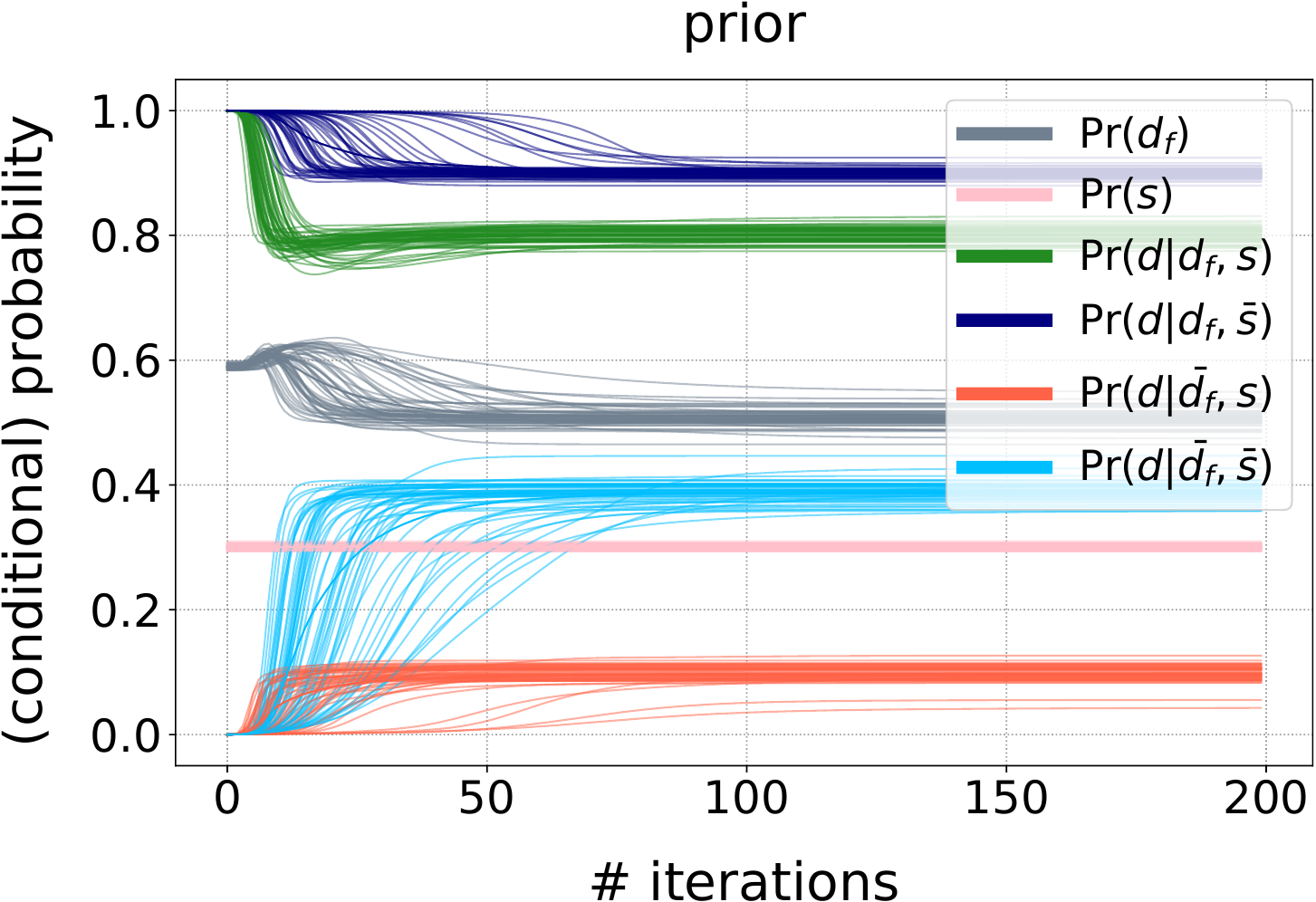}\\
    \end{tabular}
    \caption{Compare \fairpc\ initialization methods}
    \label{fig:fairpc-init}
\end{figure*}

\paragraph{Initial parameters}
Table~\ref{tab:synthetic-ll} also shows the test log-likelihoods of \fairpc\ with random initialization of parameters or with prior knowledge (i.e.\ $D=D_f$). We can see that \fairpc-prior achieves equal or slightly better results than \fairpc-rand, but the difference is not significant.
Figure~\ref{fig:fairpc-init} compares the initialization methods of \fairpc based on the conditional probability tables among $D$, $D_f$, and $S$ w.r.t.\ the number of iterations.
Each line in the plot is a single run with a certain random seed and number of non-sensitive features. From the plot, it is clear that prior knowledge outperforms random initialization in terms of convergence rate as well as the values they converge to.
The probability tables of experiments initialized from prior knowledge converge very close to the true distribution in Section~\ref{sec:exp}. However, random initialization has more variance, and some results are far from the true distribution, For example, the $P(D=1\given D_f=0, S=0)$ of several runs (light blue lines) are close to 1.0, far from 0.4; the $P(D=1\given D_f=1, S=0)$ of several runs (dark blue lines) are close to 0.4, far from 0.9. In these runs, the values of $D_f$ are switched when $S=0$.

\subsection{Learning With Missing Values}
\begin{figure}[!ht]
    \centering
    \begin{tabular}{cc}
    \includegraphics[width=0.47\columnwidth]{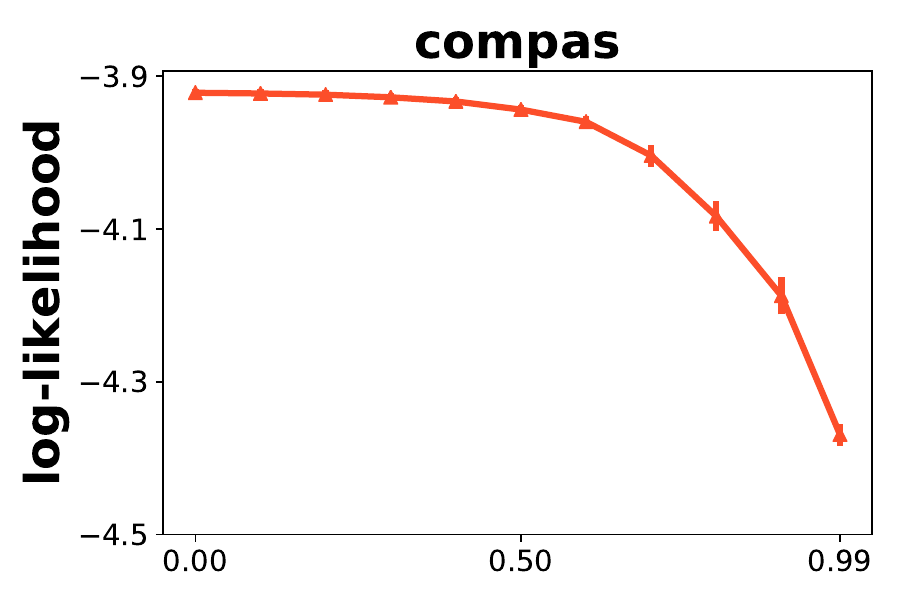} & 
    \includegraphics[width=0.47\columnwidth]{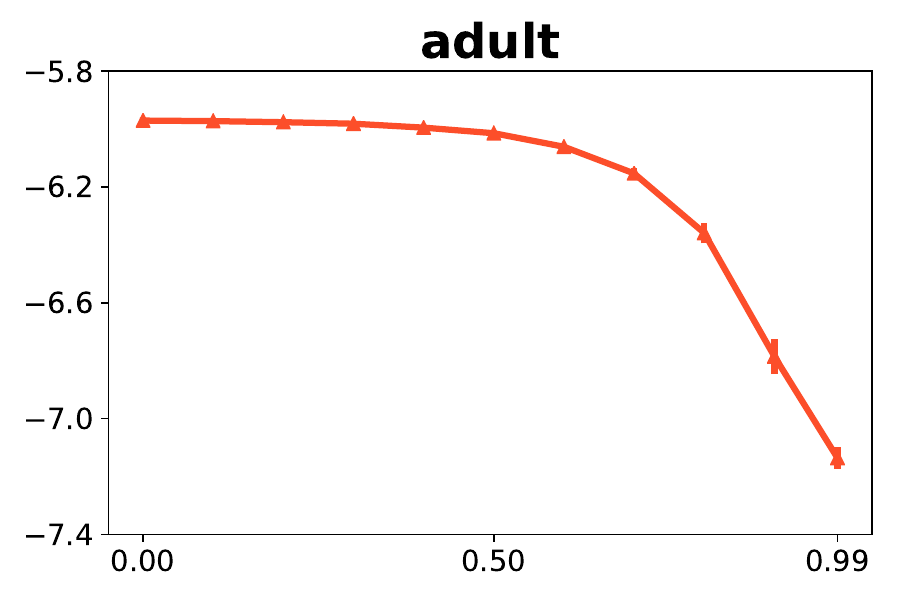} \\
    \includegraphics[width=0.47\columnwidth]{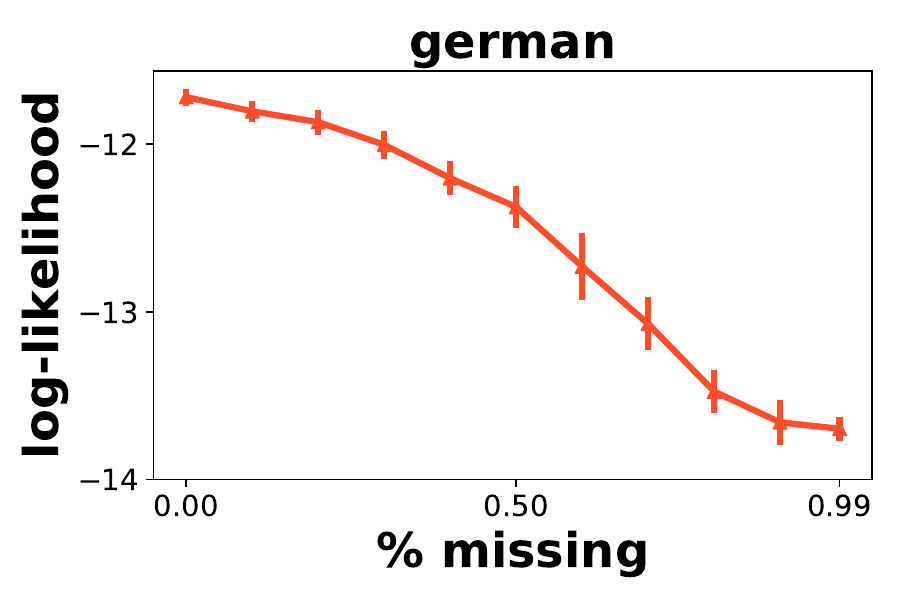} &
    \includegraphics[width=0.47\columnwidth]{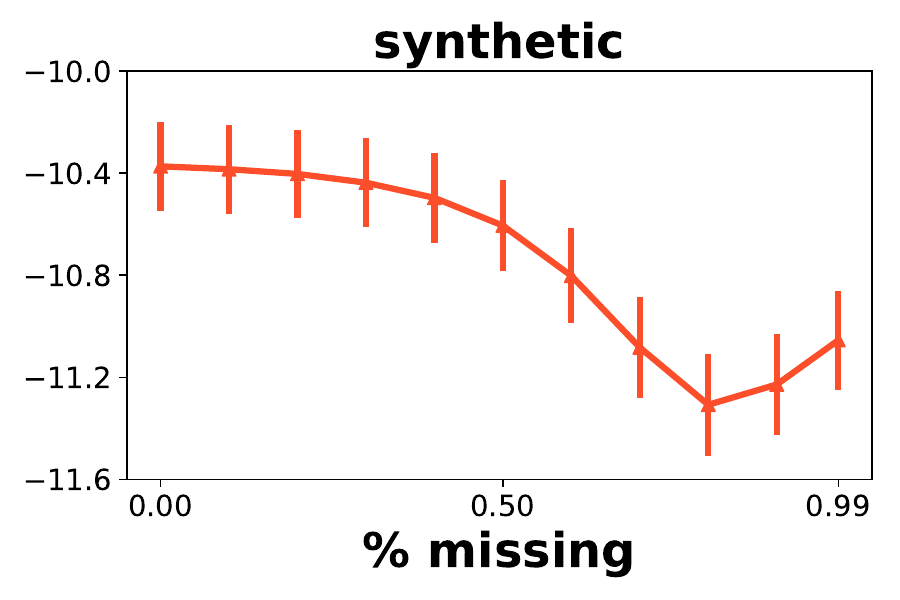}\\
    \end{tabular}
    \caption{\label{fig:missing}Test log-likelihood under different missingness percentages on real world and synthetic datasets.}
\end{figure}

As described in Section~\ref{sec:param-learn}, if the training data has some missing values (in addition to the latent decision variable), \fairpc\ parameter learning method still applies without change of the algorithm. Table~\ref{fig:missing} shows the test log-likelihoods given missing values at training time, with missing percentage ranging from 0\% to 99\%. 
We adopt missing completely at random (MCAR) missingness mechanism and fix the circuit structure to the ones learned in Section~\ref{sec:exp}. We only compare the density estimation performance here, comparing prediction performance as well as their fairness implications under different missingness is left as future work.

\begin{table*}[!ht]%
  \centering
    \caption{\label{tab:exp-realworld} Comparison on real world datasets}
  \scalebox{0.85}{
  \begin{tabular}{llll rrrr}
    \toprule
    Name & \# Features & \# Samples & Method & Log-likelihood & Accuracy  & F1-score & Discrimination\\ 
    \midrule
COMPAS & 7 & 60843 & \fairpc    &\textbf{-3.919}& \textbf{0.881}    & 0.868         & \textbf{0.009}    \\
       &&          & \latentnb  &-4.210         & \textbf{0.881}    &\textbf{0.897} & 0.036             \\
       &&          & \nlatent   &-3.922         & 0.877             & 0.723         & 0.024             \\
       &&          & \twonb     &-4.228         & 0.879             & 0.808         & 0.057             \\\cmidrule(r){4-8}
       &&          & \fairclass &N/A            & 0.878$\uparrow$     & 0.699$\uparrow$ & 0.007$\downarrow$   \\
       &&          & \fairreduct&N/A            & 0.882$\downarrow$   & 0.769$\uparrow$ & 0.011$\uparrow$	    \\
       &&          & \reweight  &N/A            & 0.882$\downarrow$   & 0.764$\uparrow$ & 0.049$\uparrow$	    \\
       \midrule
Adult  &13 &32561   & \fairpc   &\textbf{-5.962}& 0.822             & 0.674         & \textbf{0.028}    \\
       &&           & \latentnb &-6.515         & 0.634             & \textbf{0.761}& 0.132             \\
       &&           & \nlatent  &-5.980         & \textbf{0.831}    & 0.649         & 0.084             \\
       &&           & \twonb    &-6.764         & 0.823             & 0.725         & 0.205             \\ \cmidrule(r){4-8}
       &&           & \fairclass&N/A            & 0.786$\uparrow$     & 0.377$\uparrow$ & 0.009$\downarrow$	\\
       &&           & \fairreduct&N/A           & 0.821$\uparrow$     & 0.625$\uparrow$ & 0.019$\downarrow$	\\
       &&           & \reweight &N/A            & 0.803$\uparrow$     & 0.709$\downarrow$& 0.007$\downarrow$   \\
      \midrule
German & 21 &1000   & \fairpc   &\textbf{-11.422}   & 0.647         & 0.641            & 0.056              \\
      &&           & \latentnb &-11.999        & 0.499             & 0.476             & 0.081             \\
      &&           & \nlatent  &-11.454        & 0.680              & 0.663             & \textbf{0.050}   \\
      &&           & \twonb    &-12.207        & \textbf{0.683}    & \textbf{0.665}    & 0.064             \\ \cmidrule(r){4-8}
      &&           & \fairclass&N/A            & 0.711$\downarrow$    & 0.684$\downarrow$   & 0.034$\downarrow$  \\
      &&           & \fairreduct&N/A           & 0.706$\downarrow$   & 0.679$\downarrow$   & 0.071$\uparrow$	    \\
      &&           & \reweight&N/A             & 0.715$\downarrow$   & 0.684$\downarrow$   & 0.038$\downarrow$	 \\
\bottomrule
  \end{tabular}}
\end{table*}

\begin{table*}[t]
\centering
\caption{Pairwise Wilcoxon test p-values for test log-likelihoods}
\label{tab:ll-pvalue}
\scalebox{0.85}{
\begin{tabular}{lrrrrrr}
\toprule
       & \twonb & \twonb & \twonb & \latentnb & \latentnb  & \nlatent  \\
       & \latentnb & \nlatent & \fairpc &  \nlatent &  \fairpc & \fairpc \\
       \midrule
COMPAS & \textbf{9.08E-02}  & 0.00E+00    & 0.00E+00    & 0.00E+00    & 0.00E+00    & 1.42E-302     \\
Adult  & 0.00E+00  & 0.00E+00   & 0.00E+00    & 0.00E+00    & 0.00E+00   & 0.00E+00    \\
German & 5.53E-08  & 4.27E-37    & 1.53E-36    & 6.05E-19    & 9.03E-23    & \textbf{3.34E-01 }   \\\bottomrule
\end{tabular}}
\end{table*}
\begin{table*}[t]
\centering
\caption{Pairwise McNemar's test p-values for test prediction accuracy}
\label{tab:predict-pvalue}
\scalebox{0.85}{
\begin{tabular}{lrrrrrrr}
\toprule
       & \twonb          & \twonb           & \twonb          & \twonb          & \twonb             & \twonb             & \latentnb\\
       & \latentnb       & \nlatent         & \fairpc         & \fairreduct     & \reweight          & \fairclass         & \nlatent    \\\cmidrule(r){2-8}
COMPAS & \textbf{7.031E-02}       & \textbf{3.205E-01}        & \textbf{7.049E-02}       & \textbf{6.489E-02}       & \textbf{7.339E-02}          & \textbf{5.207E-01}          & \textbf{3.497E-02}         \\
Adult  & 0.000E+00       & 1.438E-05        & \textbf{3.830E-01}       & \textbf{2.099E-01}       & 2.516E-30          & 2.431E-50          & 0.000E+00         \\
German & 1.810E-17       & \textbf{8.299E-01}        & 2.444E-06       & \textbf{3.347E-02}       & 4.678E-03          & 8.730E-03          & 3.158E-15         \\ \midrule
       & \latentnb       & \latentnb        & \latentnb       & \latentnb       & \nlatent           & \nlatent           & \nlatent          \\
       & \fairpc         & \fairreduct     & \reweight        & \fairclass      & \fairpc            & \fairreduct        & \reweight         \\ \cmidrule(r){2-8}
COMPAS & \textbf{9.547E-01}       & \textbf{4.921E-01}        & \textbf{5.628E-01}       & \textbf{8.014E-02}       & \textbf{2.060E-02}          & 6.854E-05          & 1.009E-04         \\
Adult  & 0.000E+00       & 0.000E+00        & 0.000E+00       & 0.000E+00       & 2.775E-20          & 1.486E-21          & 2.101E-60         \\
German & 1.904E-08       & 8.699E-19        & 1.094E-19       & 2.806E-20       & 1.688E-06          & \textbf{4.108E-02}          & 7.096E-03         \\\midrule
       & \nlatent        & \fairpc          & \fairpc         & \fairpc         & \fairreduct        & \fairreduct        & \reweight         \\
       & \fairclass      & \fairreduct      & \reweight       & \fairclass      & \reweight          & \fairclass         & \fairclass        \\ \cmidrule(r){2-8}
COMPAS & \textbf{7.791E-01}     & \textbf{4.710E-01}        & \textbf{5.478E-01}       & \textbf{6.730E-02}       & \textbf{6.392E-01}          & 7.083E-04          & 8.349E-04         \\
Adult  & 1.017E-104      & \textbf{5.240E-01}        & 8.664E-34       & 1.560E-59       & 5.095E-29          & 1.574E-65          & 1.338E-11         \\
German & \textbf{1.911E-02}       & 4.913E-09        & 5.139E-10       & 6.094E-10       & \textbf{1.060E-01}          & \textbf{4.458E-01}          & \textbf{6.115E-01}         \\\bottomrule
\end{tabular}}
\end{table*}

\begin{table*}[ht]
  \centering
  \caption{Comparison of log-likelihoods on synthetic datasets}
  \label{tab:synthetic-ll}
  \scalebox{0.85}{
\begin{tabular}{lrrrrrrrrrrr}

\toprule
        & \multicolumn{11}{c}{\# non-sensitive variables}\\
        & 10           & 11           & 12           & 13           & 14           & 15            & 16            & 17            & 18            & 19            & 20            \\\midrule
True     & -6.975  & -7.468  & -8.091  & -8.573  & -9.149  & -9.808   & -10.210  & -11.119  & -11.296  & -11.718  & -12.476  \\
\latentnb  & -7.361  & -8.068  & -8.614  & -9.146  & -9.832  & -10.677  & -10.989  & -11.626  & -12.524  & -13.003  & -13.592  \\
\nlatent & -7.059  & -7.612  & -8.224  & -8.717  & -9.392  & -10.057  & -10.537  & -11.408  & -11.680  & -12.207  & -12.971  \\
\fairpc-rand    & -7.024  & -7.540  & -8.166  & -8.644  & -9.281  & -9.920   & -10.407  & -11.269  & -11.513  & -11.973  & -12.755  \\
\fairpc-prior   & -7.022  & -7.540  & -8.163  & -8.644  & -9.276  & -9.920   & -10.405  & -11.269  & -11.513  & -11.973  & -12.755  \\
\bottomrule
\end{tabular}}
\end{table*}
\end{document}